%% file: main.tex
\documentclass[10pt,twocolumn,letterpaper]{article}

\usepackage{cvpr}              

\input{preamble}

\definecolor{cvprblue}{rgb}{0.21,0.49,0.74}
\definecolor{linkcolor}{HTML}{ED1C24}
\usepackage[pagebackref,breaklinks,colorlinks,allcolors=cvprblue]{hyperref}
\hypersetup{
    colorlinks=true,
    urlcolor=magenta
}

\def\paperID{***} 
\def\confName{CVPR}
\def\confYear{2026}

\title{NuWa: Deriving Lightweight Class-Specific Vision Transformers\\for Edge Devices}
\author{Ziteng Wei$^{1}$ \hfill
Qiang He$^{1,2,}$\thanks{Corresponding author.} \hfill
Bing Li$^{1}$ \hfill
Feifei Chen$^{3,*}$ \hfill
Hai Jin$^{1}$ \hfill
Yun Yang$^{2}$ \\
$^1$National Engineering Research Center for Big Data Technology and System, \\
Services Computing Technology and System Lab, Cluster and Grid Computing Lab, \\
Huazhong University of Science and Technology\\
$^2$ Swinburne University of Technology
$^3$ Deakin University \quad
\\
{\texttt{\small 
\{weiziteng,hqiang,lbing,hjin\}@hust.edu.cn~ \hfill 
feifei.chen@deakin.edu.au~ \hfill
yyang@swin.edu.au
}}}

\begin{document}
\maketitle
\input{sec/0_abstract}    
\input{sec/1_intro}

\input{sec/2_related_work}

\input{sec/3_methodology}

\input{sec/4_experiment}

\input{sec/5_conclusion}
\clearpage
\input{sec/6_acknowledgment}
{
    \small
    \bibliographystyle{ieeenat_fullname}
    \bibliography{bibliography}
}

\input{sec/X_suppl}

\end{document}

%% file: preamble.tex
\usepackage[skip=3pt]{caption}      
\usepackage{multirow} 
\usepackage[table,xcdraw]{xcolor}
\definecolor{upred}{HTML}{B94A37}
\definecolor{downblue}{HTML}{2F46C5}
\usepackage{amsthm}
\usepackage{algorithm}
\usepackage{algorithmic}



%% file: sec/0_abstract.tex
\begin{abstract}
Vision Transformers (ViTs) often need to be compressed for deployment on resource-constrained edge devices like drones and smart vehicles. However, existing model compression methods ignore that many edge devices only require the knowledge of specific classes for their applications. As a result, the derived all-class ViTs retain redundant knowledge and perform suboptimally on these classes. 
We discovered that simply replacing the calibration dataset with class-specific data does not suffice to address this issue, as these methods face two fundamental limitations. 
First, they overlook the existence of class-detrimental weights, which interfere with specialization, while removing them can improve class-specific performance.
Second, the diversity of target classes and resource constraints on edge devices demand numerous customized models. Existing methods are time-consuming and computationally expensive, thus unscalable. In this work, we present NuWa, a cost-efficient method that addresses these challenges by deriving small ViTs from base ViTs for edge devices with specific class requirements. 
NuWa performs self-knowledge purification to prune class-detrimental weights and efficiently derives compact ViTs through closed-form optimization.
Without post-pruning retraining, the derived edge ViTs surpass the base ViT in class-specific accuracy and accelerate inference. Comprehensive experiments demonstrate that NuWa outperforms state-of-the-art training-free pruning methods on class-specific tasks by up to 29.00\% in accuracy. Compared with the best-performing training-dependent pruning method, NuWa achieves a 33.69× pruning speedup and reduces pruning cost by up to 99.83\%, with only a 0.61\% average accuracy loss. Project Page: \url{https://github.com/CGCL-codes/NuWa}
\end{abstract}




%% file: sec/1_intro.tex
\section{Introduction}
\label{sec:intro}

Vision Transformers (ViTs)~\cite{dosovitskiy2020image} have been widely adopted to facilitate visual services like image recognition~\cite{lu2025dhvt,fixelle2025hypergraph}, object detection~\cite{yu2024spatial,wang2025object}, and instance segmentation~\cite{jain2023oneformer,ravi2025sam}. Enabling real-time visual services for edge devices like smart vehicles and drones is becoming increasingly important in improving the quality of people's everyday lives~\cite{ye2024galaxy,hao2025nazar}. However, most ViTs demand massive computation and storage resources, making deployment on resource-constrained edge devices a grand challenge~\cite{zhang2024dense,jiang2025janus}.

\begin{figure}[t]
   \centering
   \includegraphics[width=0.95\linewidth]{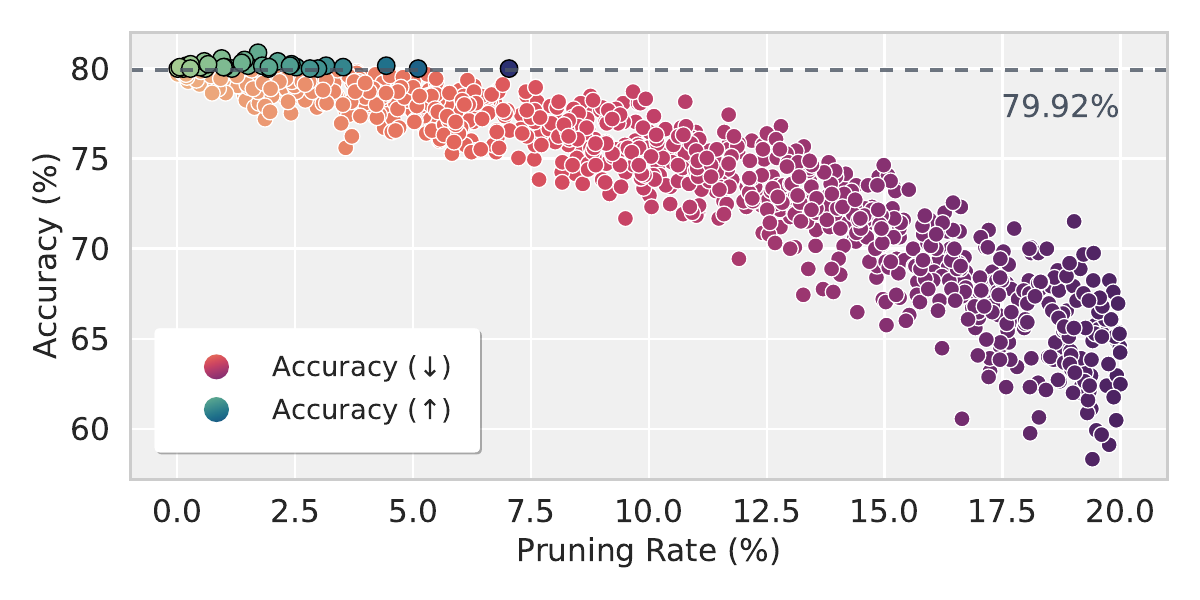}
   \caption{Pruning can sometimes improve class-specific performance. Randomly removing certain neurons from the MLP modules of DeiT-Base unexpectedly increases model accuracy on specific classes, revealing the existence of class-detrimental weights.}
   \label{fig:less_is_more}
\end{figure}

\begin{figure*}[t]
   \centering
   \includegraphics[width=0.95\linewidth]{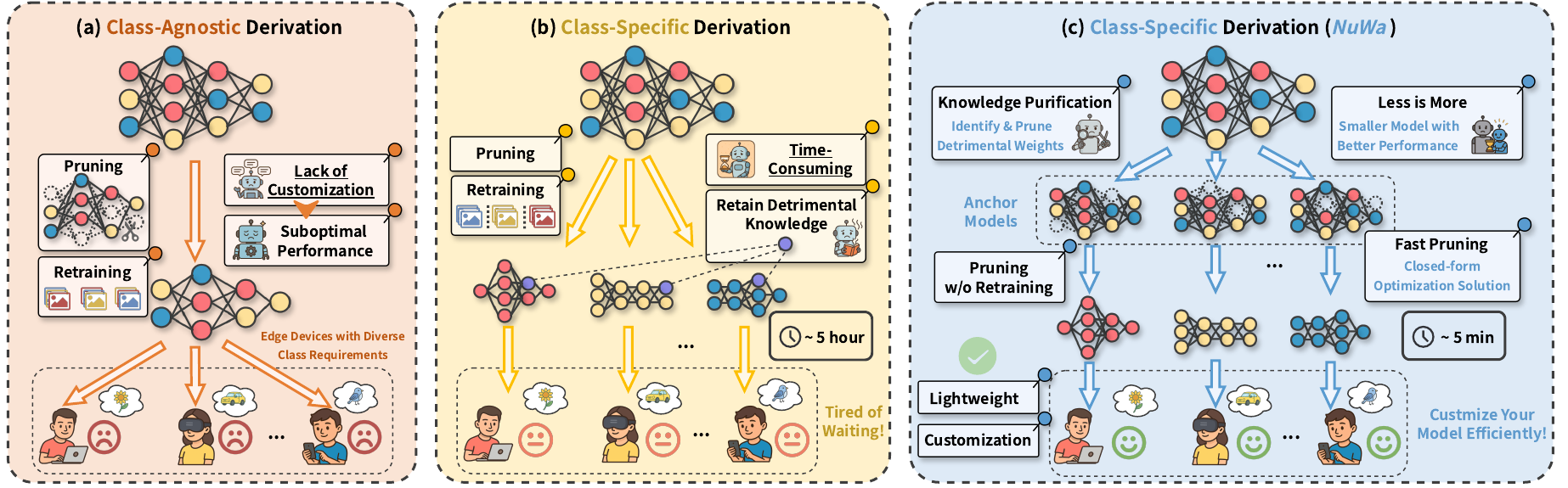}
   \caption{Comparison of model derivation settings for ViTs. (a) Class-agnostic derivation compresses the base ViT without considering class differences, lacking customization for diverse scenarios. (b) Class-specific derivation uses class-specific data for pruning and retraining but fails to remove class-detrimental knowledge and is time-consuming, limiting scalability. (c) NuWa removes class-detrimental weights and formulates pruning as closed-form optimization problems, enabling fast derivation of lightweight and customized edge ViTs.}
   \label{fig:derivation_paradigm}
   \vskip -0.1in
\end{figure*}

To tackle this challenge, many model compression methods have been proposed to adapt ViTs to edge devices~\cite{zhong2025towards,cao2025move,sun2025mdp,liao2023can}. Among them, structured pruning~\cite{cheng2024survey,fang2023depgraph-cvpr} is particularly edge-friendly due to its hardware compatibility and scalability (\cref{sec:related_works}). However, existing structured pruning methods~\cite{sun2025mdp,yang2023global,yu2023x,zhang2024dense,shen2025numerical} mainly pursue model size reduction and ignore the fact that in many scenarios, edge devices focus on specific classes and demand only part of the knowledge from the base ViT~\cite{yao2023model,zhuang2024nebula}. For example, a ViT deployed on a smart vehicle typically needs to focus on recognizing objects like pedestrians, vehicles, and traffic signs, while knowledge about flowers or birds is unnecessary. Class-irrelevant knowledge wastes model capacity, which degrades ViTs’ performance on target classes~\cite{geirhos2020shortcut}. As shown in \cref{fig:derivation_paradigm}(a), this class-agnostic derivation setting fails to provide class-specific models for edge devices, resulting in suboptimal accuracy.

\begin{figure}[t]
   \centering
   \includegraphics[width=0.95\linewidth]{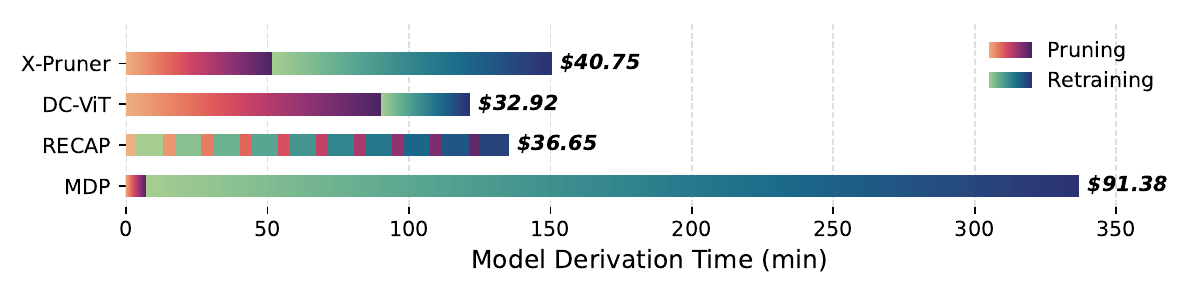}
   \caption{Existing pruning methods are time-consuming and costly. Experiments are conducted at a pruning rate of 0.50 using randomly selected 25 ImageNet classes to derive one model from DeiT-Base. Cost is based on AWS EC2 g5.48xlarge pricing.}
   \label{fig:time_consuming}
\end{figure}

Existing pruning methods can adapt to target classes by replacing the calibration dataset with class-specific data, as illustrated in \cref{fig:derivation_paradigm}(b). However, they suffer from two inherent limitations. 1) \textit{Neglect of detrimental weights}. They mainly remove unimportant weights to minimize the loss of useful knowledge during pruning. However, under the class-specific derivation setting, we observe that some class-detrimental weights are more worth removing, as pruning them can improve the model performance on target classes. We randomly select 25 classes from ImageNet-1K~\cite{russakovsky2015imagenet} and remove a subset of neurons from the Multi-Layer Perceptron (MLP) modules of DeiT-Base~\cite{touvron2021training-icml}. Interestingly, we find that pruning certain weights can increase the accuracy of DeiT-Base on these 25 classes, as shown in \cref{fig:less_is_more}. These detrimental weights cannot be captured by conventional importance metrics~\cite{han2015learning,sun2023simple,yu2018nisp,molchanov2019pruning} (\cref{sec:self_knowledge_purification}). The derived edge ViTs fail to effectively focus on the target classes. 2) \textit{High time and computation cost}. Different application scenarios involve distinct sets of target classes, and edge devices vary in their available resources~\cite{wen2023adaptivenet}. This heterogeneity calls for a large number of customized models with appropriate sizes. Existing pruning methods are time-consuming, as they require extensive search for pruning configurations and often rely on computationally-expensive retraining to recover model accuracy. These incur considerable computational and time overheads~\cite{ilhan2024resource}, as shown in \cref{fig:time_consuming}. Moreover, they lack reusable intermediate results. New target classes and pruning rates require pruning from scratch. For these reasons, existing pruning methods are impractical for large-scale edge deployment.

To address these challenges, this paper presents NuWa, a novel and cost-efficient method that can derive accurate and lightweight class-specific ViTs from base ViTs through \textit{structured pruning} without post-pruning retraining. As shown in \cref{fig:derivation_paradigm}(c), NuWa first applies \textit{self-knowledge purification} (SKP) to filter out class-detrimental knowledge from the base ViT $\mathcal{V}_B$. Specifically, NuWa freezes $\mathcal{V}_B$ and embeds learnable mask vectors and control factors into its MLP modules to construct a pruning space. Without additional regularization terms, $\mathcal{V}_B$ can leverage the original task loss $\mathcal{L}_\text{T}$ and class-specific data $\mathcal{D_S}$ to identify and prune the class-detrimental weights autonomously. SKP can be viewed as a process of discovering the “free lunch” within ViTs, as it produces a smaller yet more accurate anchor model $\mathcal{V}_A$. Since the amount of detrimental knowledge is limited, $\mathcal{V}_A$ often requires further compression to meet resource constraints. To achieve this efficiently, NuWa formulates the pruning tasks of Multi-Head Attention (MHA) and MLP modules as optimization problems and derives their \textit{closed-form solutions}, i.e., analytical solutions, to eliminate the need for retraining. Given resource constraints (\cref{sec:gflops} in supplementary material (Suppl.)) and target classes, NuWa can rapidly derive class-specific edge ViTs $\mathcal{V}_E$ with appropriate size from $\mathcal{V}_B$. 
Our main contributions are as follows:
\begin{itemize}[leftmargin=0.2in, topsep=0.01in, itemsep=0.0in]
    \item We reveal the existence of class-detrimental knowledge under the class-specific model derivation setting and propose self-knowledge purification (SKP) to identify and prune such detrimental weights.
    \item We formulate the pruning of ViT’s MHA and MLP modules as optimization problems and derive their closed-form solutions, providing an efficient and principled approach for fast model derivation.
    \item We present NuWa, which leverages these innovative strategies to derive small and accurate class-specific ViTs. To the best of our knowledge, it is the first method for deriving class-specific ViTs.
    \item Extensive experiments with six models on four datasets show that NuWa effectively derives lightweight class-specific ViTs, outperforming state-of-the-art training-free and training-dependent structured pruning methods in class-specific performance and efficiency.
\end{itemize}

%% file: sec/2_related_work.tex
\section{Related Works}
\label{sec:related_works}

\textbf{Model Compression.} Various methods have been developed to derive lightweight ViTs, including low-bit quantization~\cite{li2022q,liu2023oscillation,choi2025gradq,zhong2025towards}, knowledge distillation~\cite{hao2024one,yang2024vitkd,cao2025move}, as well as structured~\cite{zhang2024dense,sun2025mdp} and unstructured~\cite{chen2021chasing,liao2023can} pruning. However, some of these methods suffer notable drawbacks (\cref{sec:limitations_of_compression_methods} in Suppl.). Low-bit quantization and unstructured pruning often require specialized infrastructure for inference, which limits their applicability to framework-diverse edge devices~\cite{yang2024laco,cheng2024survey}. Knowledge distillation does not always provide a suitable student model with an appropriate size for initialization, and training a new one from scratch often incurs excessive computational cost~\cite{yang2024laco,he2025dakd}. In contrast, structured pruning produces regularly shaped models that are easy to deploy on diverse edge devices and can transfer knowledge from the parameter space of the base ViT for initialization~\cite{an2024fluctuation,ling2024slimgpt}. It offers the generality, feasibility, and portability needed for edge ViT derivation, thus serving as the foundation of NuWa. It is worth noting that NuWa can be integrated with other compression methods for further model compression~\cite{frantar2022optimal,muralidharan2024compact}.

\textbf{Structured Pruning for Transformers}. There are two types of structured pruning methods, i.e., training-free and training-dependent pruning.
1) \textit{Training-free pruning} methods~\cite{zhou2022training,han2015learning,sun2023simple,shen2025numerical} prune weights based on importance metrics such as weight magnitude, activation, or gradient (\cref{sec:metric} in Suppl.). Some methods~\cite{kim2020neuron,ling2024slimgpt,shen2025numerical} further compute compensation weights to align the pruned and original models, mitigating accuracy loss. However, their objective is to make the pruned models mimic the base model, ignoring the existence of class-detrimental weights (\cref{fig:less_is_more}) under the class-specific model derivation setting. They cannot inject class-specific knowledge through training and inherently assume that pruning always leads to accuracy degradation~\cite{tran2022pruning}. As a result, the derived models can never surpass the base model on target classes. In contrast, NuWa removes class-detrimental weights through SKP (\cref{sec:self_knowledge_purification}) and elevates the performance of derived models on target classes during pruning.
2) \textit{Training-dependent pruning} methods~\cite{yu2023x,zhang2024dense,ilhan2024resource,sun2025mdp} also compute importance scores or train sparse masks to guide pruning. Since they introduce retraining after or during pruning to recover accuracy, they can leverage class-specific data to improve the model's focus on target classes. However, they still fail to remove class-detrimental weights. More importantly, retraining often takes a long time and incurs excessive computational overhead. This makes them impractical for large-scale model deployment on diverse edge devices with different class requirements and resource constraints~\cite{wen2023adaptivenet,cai2019once,scherer2024deeploy,lu2025demystifying}. Instead, NuWa computes pruned weights directly, enabling effective and fast model derivation.

%% file: sec/3_methodology.tex
\begin{figure}[t]
   \centering
   \includegraphics[width=0.91\linewidth]{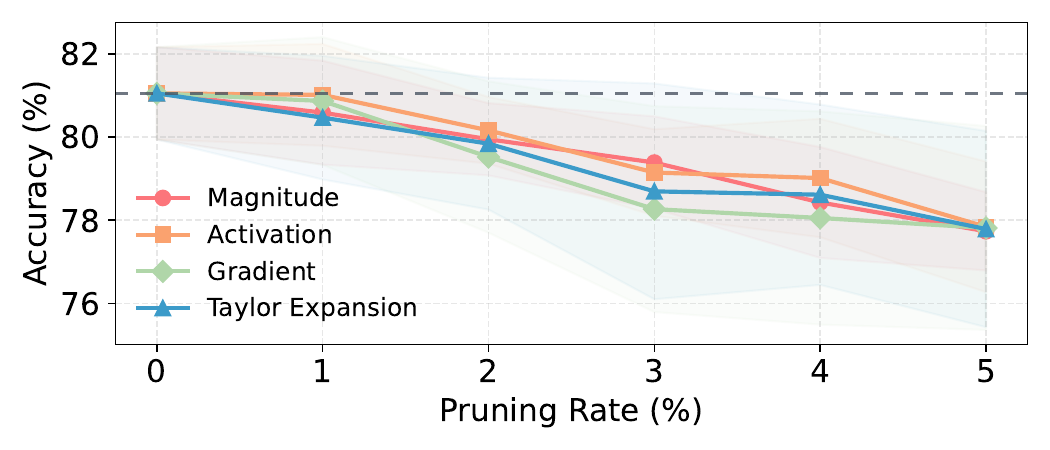}
   \caption{Existing importance metrics fail to improve class-specific accuracy through pruning. The solid lines with standard deviation bands represent the mean accuracy of pruned DeiT-Base on three random sub-tasks ($|\mathcal{S}|$=25) across different pruning rates.}
   \label{fig:importance_metric}
\end{figure}

\begin{figure*}[t]
   \centering
   \includegraphics[width=0.95\linewidth]{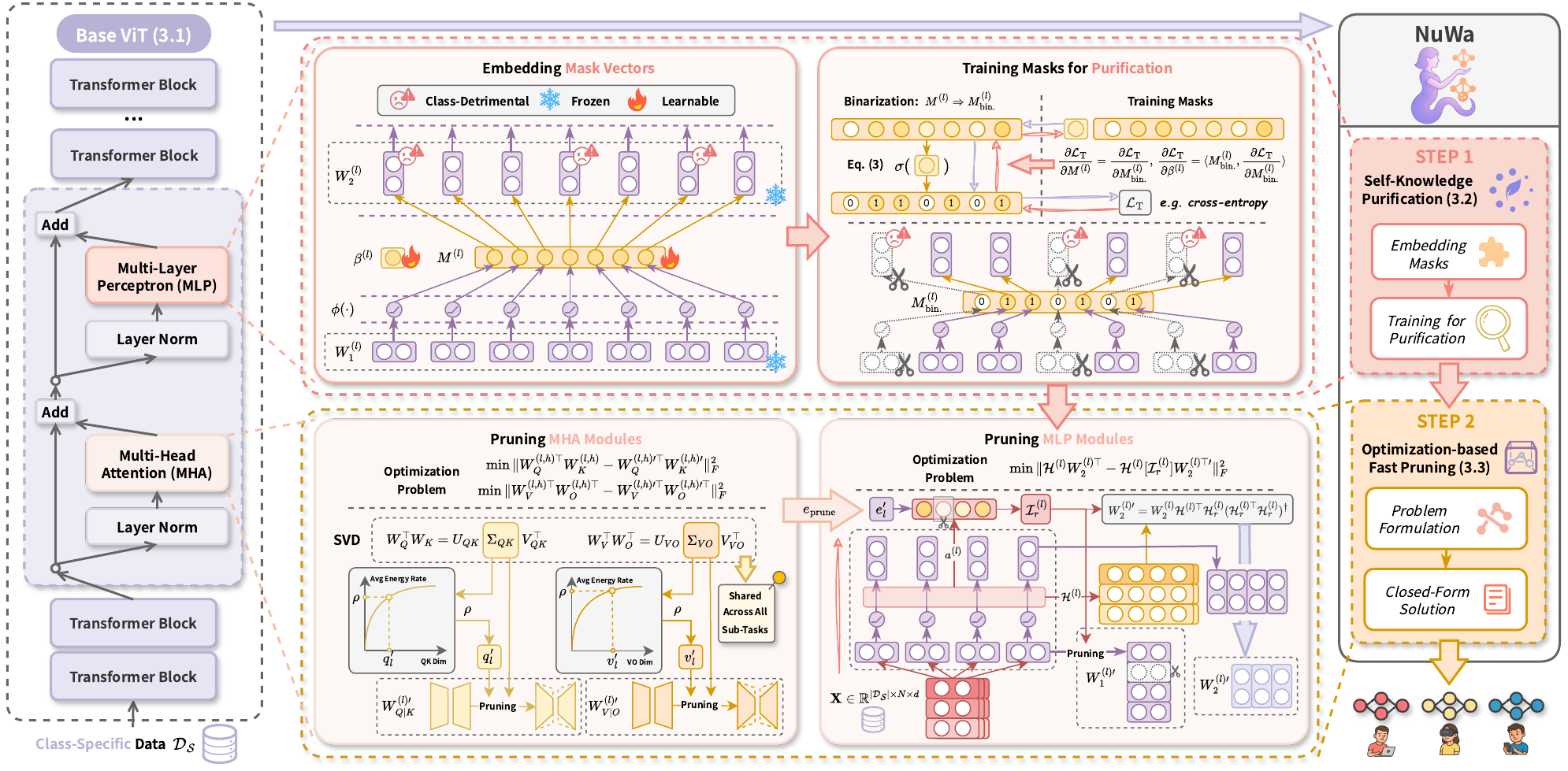}
   \caption{Overview of NuWa. 
   1) \textit{Self-Knowledge Purification} (SKP): learning binarized masks to identify and prune class-detrimental weights. 
   2) \textit{Optimization-based Fast Pruning} (OFP): pruning MHA and MLP modules through closed-form optimization.}
   \label{fig:overview}
   \vskip -0.2in
\end{figure*}

\section{Methodology}
\label{sec:methodology}

This section presents NuWa in detail, starting with the notations and preliminaries (\cref{sec:preliminaries}). Then, it presents the details of the two key components of NuWa, i.e., Self-Knowledge Purification (SKP) and Optimization-based Fast Pruning (OFP). SKP constructs a pruning space that enables the model to autonomously identify and prune class-detrimental weights, aiming to improve model performance on target classes and provide an anchor model for subsequent optimization (\cref{sec:self_knowledge_purification}). Next, OFP performs fast pruning with closed-form optimization to efficiently derive lightweight ViTs from the anchor model (\cref{sec:optimization_based_pruning}). An overview of NuWa is illustrated in \cref{fig:overview}.

\subsection{Notations and Preliminaries}
\label{sec:preliminaries}

\textbf{Vision Transformer}. A ViT~\cite{dosovitskiy2020image} consists of a patch embedding layer, a series of transformer blocks, and a classifier. Each transformer block includes an MHA  and MLP module. Given an input $\mathbf{X}\in\mathbb{R}^{\smash{N\times d}}$ with $N$ d-dimensional patch tokens, each head in the MHA module computes an attention $\mathcal{A}$ independently before the results are summed:
\begin{equation}
\label{eq:attention}
\begin{aligned}
    &\hspace{-0.4em}\mathcal{A}^{(l,h)}=\text{Head}^{(l,h)}(\mathbf{X})=\text{Softmax}\big(\frac{QK^\top}{\sqrt{q_l}}\big)V{W^{(l,h)\top}_O}\\
    &\hspace{-0.4em}=\text{Softmax}\big(\frac{\mathbf{X}{W^{(l,h)\top}_Q} W^{(l,h)}_K\mathbf{X}^\top}{\sqrt{q_l}}\big)\mathbf{X}{W^{(l,h)\top}_V}{W^{(l,h)\top}_O}
\end{aligned}
\end{equation}
where $W^{\smash{(l,h)}}_{\smash{Q|K}}\in\mathbb{R}^{\smash{q_l\times d}}$, $W^{\smash{(l,h)}}_{\smash{V}}\in\mathbb{R}^{\smash{v_l\times d}}$ and $W^{\smash{(l,h)}}_{\smash{O}}\in\mathbb{R}^{\smash{d\times v_l}}$ denote the weights of the $h$-th head in the $l$-th block. Bias terms are omitted here for clarity. Given the MHA outputs, also denoted by $X$ for simplicity, the MLP module weighted-sums the outputs of all $e_l$ neurons:
\begin{equation}
\label{eq:multi_layer_perceptron}
    \text{MLP}^{(l)}(\mathbf{X})=\sum_{i=1}^{e_l}\phi(\mathbf{X}W_1^{(l)}[i])\otimes {{W^{(l)\top}_2}[i]}
\end{equation}
where $W_1^{\smash{(l)}}\in\mathbb{R}^{\smash{e_l\times d}}$ and $W_2^{\smash{(l)}}\in\mathbb{R}^{\smash{d\times e_l}}$ are the weights of the MLP module in the $l$-th block, $\phi(\cdot)$ is a nonlinear activation (e.g., GELU), $W[i]$ represent the $i$-th row vectors of $W$, and $\otimes$ is the outer product.

\textbf{Sub-Tasks}. Let $\mathcal{Y} = \{y_1, \cdots, y_C\}$ denote the class set of the pre-training dataset for the base ViT $\mathcal{V}_B$. We aim to derive an edge ViT from $\mathcal{V}_B$, denoted as $\mathcal{V}_E$, that performs a \textit{sub-task} of $\mathcal{V}_B$ to perceive a subset of $\mathcal{Y}$, denoted as $\mathcal{S} \subset \mathcal{Y}$. The corresponding class-specific data is denoted as $\mathcal{D_S}$.

\subsection{Self-Knowledge Purification}
\label{sec:self_knowledge_purification}

As illustrated in \cref{fig:less_is_more}, under the class-specific model derivation setting, there are class-detrimental weights, the removal of which can improve the model's performance on specific classes. However, as shown in \cref{fig:importance_metric}, existing importance evaluation metrics can not be used to identify these weights, even when calculated with class-specific data $\mathcal{D_S}$.

\textbf{Constructing Pruning Space}. Since there is no prior knowledge indicating which weights are class-detrimental or how many of them exist, we propose SKP to automatically filter class-detrimental knowledge by learning binary masks. Specifically, NuWa freezes the base ViT $\mathcal{V}_B$ and embeds a learnable mask vector $M^{\smash{(l)}}\in\mathbb{R}^{\smash{e_l}}$ and a control factor $\beta^{\smash{(l)}}\in\mathbb{R}^{\smash{1}}$ before the down-sampling weight $W^{\smash{(l)}}_2$ in each MLP. Then, NuWa introduces a binarization operation on $M^{\smash{(l)}}$ based on $\beta^{\smash{(l)}}$ to construct a pruning space for $\mathcal{V}_B$:
\begin{equation}
\label{eq:mask_binarization}
    M^{(l)}_\text{bin.}[i]=\begin{cases}
        1, & \text{if } M^{(l)}[i]\ge\text{Sel}_{\lfloor e_l\times\sigma(\beta^{(l)})\rfloor}(M^{(l)})\\
        0, & \text{otherwise}
    \end{cases}
\end{equation}
where $\text{Sel}_k(v)$ returns the $k$-th largest value in vector $v$, and $\sigma(\cdot)$ denotes the sigmoid function. The binarized mask $M^{\smash{(l)}}_\text{bin.}$ then participates in the forward propagation of the MLP module, modifying \cref{eq:multi_layer_perceptron} as follows:
\begin{equation}
\label{eq:mlp_binarized}
    \text{MLP}^{(l)}(\mathbf{X})=\sum_{i=1}^{e_l}\phi(\mathbf{X}W_1^{(l)}[i])\otimes ({M^{(l)}_\text{bin.}[i]\cdot W^{(l)\top}_2}[i])
\end{equation}
This design establishes a structured pruning space for $\mathcal{V}_B$, in which setting $M^{\smash{(l)}}_\text{bin}[i]=0$ corresponds to pruning the $i$-th row of $W^{\smash{(l)}}_1$ and the $i$-th column of $W^{\smash{(l)}}_2$. Following the function-preserving principle~\cite{chen2015net2net,hu2021lora}, $\beta^{\smash{(l)}}$ in each MLP module is initialized to 5.0 ($\sigma(5.0)\approx$1.0) to ensure consistency between the model before and after the embedding of the mask vectors. NuWa does not apply SKP to the MHA modules at this stage, as empirical observations show that performing SKP only on the MLP modules yields better results (\cref{sec:design_justification} in Suppl.).

\textbf{Knowledge Purification}. To enable the model to identify and prune class-detrimental weights through the learning of $M^{\smash{(l)}}$ and $\beta^{\smash{(l)}}$, NuWa adopts a straight-through estimator (STE)~\cite{ramanujan2020s,wu2023estimator} to address the non-differentiability of the binarization operation. Specifically, the gradient of $M^{\smash{(l)}}_\text{bin.}$, denoted as $G^{\smash{(l)}}$, is used as the surrogate gradient for $M^{\smash{(l)}}$, while the inner product $\langle M^{\smash{(l)}}_\text{bin.},G^{\smash{(l)}}\rangle$ serves as the surrogate gradient for $\beta^{\smash{(l)}}$. Subsequently, NuWa uses class-specific data $\mathcal{D_S}$ to update $\mathcal{M}=\{M^{\smash{(l)}}\}^L_{l=1}$ and $\mathcal{B}=\{\beta^{\smash{(l)}}\}^L_{l=1}$ under the supervision of the original visual task loss $\mathcal{L}_\text{T}$. Finally, NuWa prunes each MLP module of $\mathcal{V}_B$ based on $\mathcal{M}^{\smash{(l)}}_\text{bin.}$ and obtains an anchor model $\mathcal{V}_A$:
\begin{equation}
\label{eq:get_anchor_model}
    W^{(l)}_1=W^{(l)}_1[\mathcal{I}(\mathcal{M}^{(l)}_\text{bin.})], W^{(l)}_2=W^{(l)}_2[:,\mathcal{I}(\mathcal{M}^{(l)}_\text{bin.})]
\end{equation}
\begin{equation}
\label{eq:get_pruned_idx}
    \mathcal{I}(v)=\{i\in[\text{dim}(v)]\mid v[i]\neq0\}
\end{equation}
Compared with random pruning (\cref{fig:less_is_more}), the anchor model achieves a higher pruning rate and greater class-specific performance improvement (\cref{fig:training_free_comparison}). Since SKP updates no more than 0.05\% of the total parameters in $\mathcal{V}_B$ and we observe that a smaller batch size enables $\mathcal{V}_B$ to perform a more thorough exploration in the pruning space (\cref{fig:batch_size}), the process of obtaining the anchor model is highly efficient.

\subsection{Optimization-based Fast Pruning}
\label{sec:optimization_based_pruning}

Since $\mathcal{V}_A$ may not be able to fully satisfy the resource constraints of edge devices, NuWa formulates the pruning of MHA and MLP modules as optimization problems to derive their closed-form optimal solutions for further pruning.

\textbf{Pruning MHA}. NuWa prunes the query–key (QK:$q$) and value–output (VO:$v$) dimensions of the MHA modules. This design is motivated by two considerations.
First, pruning these fine-grained dimensions provides a higher compression potential than pruning entire attention heads. Second, as shown in \cref{eq:attention}, the QK and VO dimensions serve as the intermediate dimensions of two weight multiplications (i.e., $W_{QK}={W}_Q^{\smash{\top}} W_K,W_{VO}={W}_V^{\smash{\top}}W_O^{\smash{\top}}\in\mathbb{R}^{\smash{d\times d}}$) and are independent of input features $\mathbf{X}$. This allows the pruning results of these dimensions to be shared across all sub-tasks, which effectively reduces redundant computation during large-scale deployment. 
Take QK dimension as an example, NuWa formulates the pruning of MHA modules as an optimization problem:
\begin{equation}
\label{eq:mha_pruning_problem}
    \min_{W_Q^{(l,h)\prime},W_K^{(l,h)\prime}}\|{W^{(l,h)\top}_Q} W^{(l,h)}_K-W^{(l,h)\prime\top}_Q W^{(l,h)\prime}_K\|^2_F
\end{equation}
where $W^{\smash{(l,h)\prime}}_{Q|K}\in\mathbb{R}^{q_l'\times d}$ denote the pruned weights with pruned QK dimension $q_l' < q_l$. This optimization problem essentially becomes a low-rank matrix approximation problem, where the objective is to find a matrix $W_{QK}^{\smash{(l,h)\prime}}$ with rank less than $q_l$, such that the Frobenius norm of its difference from $W_{QK}^{\smash{(l,h)}}$ is minimized. According to the \textit{Eckart-Young theorem} \citep{chipman2020proofs}, singular value decomposition (SVD) guarantees this minimization (\cref{sec:proof} in Suppl.). Therefore, NuWa performs MHA pruning through SVD as follows:
\begin{equation}
\label{eq:prune_mha}
\begin{aligned}
&W_Q^{(l,h)\prime} = (U^{(l,h)}_{QK}[:,:q_l']\Sigma^{(l,h)}_{QK}[:q_l',:q_l'])^\top\times\sqrt{{q_l'}/{q_l}}\\
&W_{V}^{(l,h)\prime}=(U^{(l,h)}_{VO}[:,:v_l']\Sigma^{(l,h)}_{VO}[:v_l',:v_l'])^\top\\
&W_{K}^{(l,h)\prime}=V^{(l,h)}_{QK}[:,:q_l']^\top,~ W_{O}^{(l,h)'}=V^{(l,h)}_{VO}[:,:v_l']
\end{aligned}
\end{equation}
where $U_{QK}\Sigma_{QK} V_{QK}^{\smash{\top}}$=$W_{QK}$ and $U_{VO}\Sigma_{VO} V_{VO}^{\smash{\top}}$=$W_{VO}$. NuWa employs a hyperparameter $\rho$, which represents the ratio of retained singular value energy to the total energy, to adaptively determine the pruned dimensions $q_l'$ and $v_l'$. Take the QK dimension as an example. Under the constraint of $\rho$, NuWa prunes the $q_l$ until $q_l'$ satisfies:
\begin{equation}
\label{eq:retained_energy_rate}
\frac1{H_l}\sum_{h=1}^{H_l}\left({\sum_{i=1}^{q_l'-1}{\sigma_{QK}^{(l,h,i)}}^2}/{\sum_{i=1}^{q_l}{\sigma_{QK}^{(l,h,i)}}^2}\right)<\rho
\end{equation}
where $\sigma_{QK}^{\smash{(l,h,i)}}$ denotes the $i$-th singular value of $W_{QK}^{\smash{(l,h)}}$. By adjusting $\rho$, NuWa can balance the pruning intensity of MHA and MLP modules.

\textbf{Pruning MLP}. After pruning the MHA, NuWa prunes the intermediate dimensions of the MLP module to achieve the desired overall pruning rate $\alpha$. 
To avoid excessive pruning in certain blocks, which may harm model performance, and to improve the inference efficiency~\cite{xia2023sheared} of the derived $\mathcal{V}_E$, NuWa prunes all MLP modules of $\mathcal{V}_A$ to similar sizes whenever possible (\cref{sec:design_justification} in Suppl.). The pruned intermediate dimension $e_l'$ of each block is determined as follows:
\begin{equation}
\label{eq:block_uniform_mlp_pruning}
e_l'= \min(e_l,\lfloor(\sum_{i=1}^Le_i-e_\text{prune})/L\rfloor)
\end{equation}
where $e_\text{prune}$ denotes the total number of neurons to be pruned across all MLP modules. After determining $e_l'$, NuWa performs a forward propagation of $\mathcal{V}_A$ on $\mathcal{D_S}$ to obtain the mean activation value $a^{\smash{(l)}}_i$ of each neuron in the MLP across all patches. Meanwhile, it samples $K$ images to extract their block-wise activation features $\mathcal{H}^{\smash{(l)}}\in\mathbb{R}^{\smash{(KN)\times e_l}}$. NuWa retains the top $e_l'$ neurons with the highest activation values in each MLP, whose indices are denoted by $\mathcal{I}^{\smash{(l)}}_r$. It then solves the following optimization problem:
\begin{equation}
\label{eq:mlp_pruning_problem}
    \min_{W_2^{(l)\prime}}\|\mathcal{H}^{(l)}W_2^{(l)\top}-\mathcal{H}^{(l)}[\mathcal{I}^{(l)}_r]W_2^{(l)\prime\top}\|^2_F
\end{equation}
NuWa derives a closed-form solution to the above optimization problem to minimize the knowledge loss caused by MLP pruning (\cref{sec:proof} in Suppl.).
Based on this solution, the pruning process for the MLP modules is given by:
\begin{equation}
\label{eq:prune_mlp}
\begin{aligned}
& W_1^{(l)\prime}=W_1^{(l)}[\mathcal{I}^{\smash{(l)}}_r]\in\mathbb{R}^{e_l'\times d}\\
& W_2^{(l)\prime}=W_2^{(l)}\mathcal{H}^{(l)\top}\mathcal{H}^{(l)}_r(\mathcal{H}^{(l)\top}_r\mathcal{H}^{(l)}_r)^{\dagger}\in\mathbb{R}^{d\times e_l'}
\end{aligned}
\end{equation}
where $\mathcal{H}^{\smash{(l)}}_r=\mathcal{H}^{\smash{(l)}}[\mathcal{I}^{\smash{(l)}}_r]\in\mathbb{R}^{\smash{(KN)\times e_l'}}$. Based on the closed-form solutions in \cref{eq:prune_mha} and \cref{eq:prune_mlp}, NuWa can efficiently prune $\mathcal{V}_A$ to derive the class-specific edge ViT $\mathcal{V}_E$.


%% file: sec/4_experiment.tex
\begin{table*}[t]
\centering 
\caption{Comparison between NuWa and training-dependent pruning baselines, with the best and second-best accuracies highlighted in bold and underlined, respectively. Subscripts indicate improvements over DeiT-Base. “(FT)” denotes fine-tuning on $\mathcal{D_S}$.}
\label{tab:training_dependent_comparison}
\begin{scriptsize}
\renewcommand{\arraystretch}{1.1}
\begin{tabular}{l||ccc|c||ccc|c||ccc|c}
\toprule
\textbf{Method} & $\mathcal{S}_1$/10 & $\mathcal{S}_2$/10 & $\mathcal{S}_3$/10 & \textit{Avg.} & $\mathcal{S}_4$/25 & $\mathcal{S}_5$/25 & $\mathcal{S}_6$/25 & \textit{Avg.} & $\mathcal{S}_7$/50 & $\mathcal{S}_8$/50 & $\mathcal{S}_9$/50 & \textit{Avg.}\\

\hline\hline
\textbf{DeiT-Base} & 71.20 & 81.60 & 83.00 & 78.60 & 79.92 & 82.56 & 80.67 & 81.05 & 80.48 & 79.64 & 84.96 & 81.69\\
\rowcolor[gray]{0.95}
\textbf{DeiT-Base (FT)} & 96.60 & 99.00 & 97.80 & 97.80 \textsubscript{\textcolor{upred}{$\uparrow$19.20}} & 96.40 & 96.96 & 96.96 & 96.77 \textsubscript{\textcolor{upred}{$\uparrow$15.72}} & 95.00 & 93.36 & 94.68 & 94.35 \textsubscript{\textcolor{upred}{$\uparrow$12.66}}\\

\hline\hline\rowcolor[HTML]{FFF2E0}
\textbf{Anchor Model} & 95.80 & 98.60 & 98.60 & 97.67 \textsubscript{\textcolor{upred}{$\uparrow$19.07}} & 94.80 & 97.92 & 96.72 & 96.48 \textsubscript{\textcolor{upred}{$\uparrow$15.43}} & 95.04 & 94.20 & 94.92 & 94.72 \textsubscript{\textcolor{upred}{$\uparrow$13.03}}\\
\hline\hline
\multicolumn{13}{l}{\textit{\textcolor{gray!70}{\textbf{Pruning Rate = 0.40}}}}\\
\textbf{Random} & 94.80 & 97.00 & 97.00 & 96.30 \textsubscript{\textcolor{upred}{$\uparrow$17.70}} & 94.48 & 95.12 & 95.44 & 95.01 \textsubscript{\textcolor{upred}{$\uparrow$13.96}} & 92.64 & 91.32 & 92.88 & 92.28 \textsubscript{\textcolor{upred}{$\uparrow$10.59}}\\
\rowcolor[gray]{0.95}
\textbf{X-Pruner} & 96.20 & 98.40 & 97.40 & 97.33 \textsubscript{\textcolor{upred}{$\uparrow$18.73}} & \underline{95.44} & \textbf{97.12} & 95.76 & \underline{96.11} \textsubscript{\textcolor{upred}{$\uparrow$15.06}} & 92.88 & 92.52 & 93.64 & 93.01 \textsubscript{\textcolor{upred}{$\uparrow$11.32}}\\
\textbf{DC-ViT} & 75.80 & 95.20 & 91.00 & 87.33 \textsubscript{\textcolor{upred}{$\uparrow$8.73}} & 89.44 & 91.76 & 90.24 & 90.48 \textsubscript{\textcolor{upred}{$\uparrow$9.43}} & 88.88 & 87.96 & 88.16 & 88.33 \textsubscript{\textcolor{upred}{$\uparrow$6.64}}\\
\rowcolor[gray]{0.95}
\textbf{RECAP} & \underline{96.60} & 98.20 & \underline{98.80} & \underline{97.87} \textsubscript{\textcolor{upred}{$\uparrow$19.27}} & 95.04 & 96.72 & 95.36 & 95.71 \textsubscript{\textcolor{upred}{$\uparrow$14.66}} & 93.48 & 92.84 & 93.80 & 93.37 \textsubscript{\textcolor{upred}{$\uparrow$11.68}}\\
\textbf{MDP} & 93.80 & 97.00 & 96.40 & 95.73 \textsubscript{\textcolor{upred}{$\uparrow$17.13}} & 93.36 & 96.16 & 94.32 & 94.61 \textsubscript{\textcolor{upred}{$\uparrow$13.56}} & 92.80 & 92.12 & 93.24 & 92.72 \textsubscript{\textcolor{upred}{$\uparrow$11.03}}\\
\hline\rowcolor[HTML]{FFF2E0}
\textbf{\textit{NuWa}} & 96.00 & \underline{98.60} & 98.60 & 97.73 \textsubscript{\textcolor{upred}{$\uparrow$19.13}} & 94.40 & \underline{97.04} & \underline{96.72} & 96.05 \textsubscript{\textcolor{upred}{$\uparrow$15.00}} & \underline{93.68} & \underline{92.88} & \underline{94.44} & \underline{93.67} \textsubscript{\textcolor{upred}{$\uparrow$11.98}}\\
\rowcolor[HTML]{FFF6E2}
\textbf{\textit{NuWa (FT)}} & \textbf{97.00} & \textbf{98.80} & \textbf{98.80} & \textbf{98.20 \textsubscript{\textcolor{upred}{$\uparrow$19.60}}} & \textbf{96.16} & \underline{97.04} & \textbf{96.72} & \textbf{96.64 \textsubscript{\textcolor{upred}{$\uparrow$15.59}}} & \textbf{93.96} & \textbf{93.52} & \textbf{94.80} & \textbf{94.09 \textsubscript{\textcolor{upred}{$\uparrow$12.40}}}\\

\hline\hline
\multicolumn{13}{l}{\textit{\textcolor{gray!70}{\textbf{Pruning Rate = 0.60}}}}\\
\textbf{Random} & 88.00 & 95.00 & 92.40 & 91.80 \textsubscript{\textcolor{upred}{$\uparrow$13.20}} & 90.48 & 93.36 & 91.52 & 91.79 \textsubscript{\textcolor{upred}{$\uparrow$10.74}} & 89.48 & 87.64 & 89.64 & 88.92 \textsubscript{\textcolor{upred}{$\uparrow$7.23}}\\
\rowcolor[gray]{0.95}
\textbf{X-Pruner} & 91.20 & 95.40 & 95.60 & 94.07 \textsubscript{\textcolor{upred}{$\uparrow$15.47}} & \underline{91.58} & 93.04 & \underline{93.11} & \underline{92.58} \textsubscript{\textcolor{upred}{$\uparrow$11.53}} & \underline{90.24} & \underline{88.18} & \underline{90.07} & \underline{89.50} \textsubscript{\textcolor{upred}{$\uparrow$7.81}}\\
\textbf{DC-ViT} & 68.20 & 79.20 & 73.80 & 73.73 \textsubscript{\textcolor{downblue}{$\downarrow$4.87}} & 70.56 & 75.68 & 75.84 & 74.02 \textsubscript{\textcolor{downblue}{$\downarrow$7.02}} & 69.84 & 69.44 & 71.32 & 70.20 \textsubscript{\textcolor{downblue}{$\downarrow$11.49}}\\
\rowcolor[gray]{0.95}
\textbf{RECAP} & 89.40 & 95.40 & 93.80 & 92.87 \textsubscript{\textcolor{upred}{$\uparrow$14.27}} & 88.88 & 92.08 & 90.24 & 90.40 \textsubscript{\textcolor{upred}{$\uparrow$9.35}} & 88.60 & 87.88 & 89.16 & 88.55 \textsubscript{\textcolor{upred}{$\uparrow$6.86}}\\
\textbf{MDP} & 90.00 & 95.80 & \underline{95.80} & 93.87 \textsubscript{\textcolor{upred}{$\uparrow$15.27}} & 90.88 & \underline{94.80} & 91.52 & 92.40 \textsubscript{\textcolor{upred}{$\uparrow$11.35}} & 90.00 & 85.00 & 90.04 & 88.35 \textsubscript{\textcolor{upred}{$\uparrow$6.66}}\\
\hline\rowcolor[HTML]{FFF2E0}
\textbf{\textit{NuWa}} & \underline{91.80} & \underline{96.20} & 94.60 & \underline{94.20} \textsubscript{\textcolor{upred}{$\uparrow$15.60}} & 89.84 & 92.00 & 92.48 & 91.44 \textsubscript{\textcolor{upred}{$\uparrow$10.39}} & 85.32 & 86.20 & 85.96 & 85.83 \textsubscript{\textcolor{upred}{$\uparrow$4.14}}\\
\rowcolor[HTML]{FFF6E2}
\textbf{\textit{NuWa (FT)}} & \textbf{95.20} & \textbf{98.40} & \textbf{97.80} & \textbf{97.13 \textsubscript{\textcolor{upred}{$\uparrow$18.53}}} & \textbf{95.36} & \textbf{96.88} & \textbf{95.60} & \textbf{95.95 \textsubscript{\textcolor{upred}{$\uparrow$14.90}}} & \textbf{92.32} & \textbf{92.12} & \textbf{93.44} & \textbf{92.63\textsubscript{\textcolor{upred}{$\uparrow$10.94}}}\\
\hline
\bottomrule
\end{tabular}
\end{scriptsize}
\vskip -0.15in
\end{table*}

\begin{figure}[t]
   \centering
   \includegraphics[width=0.95\linewidth]{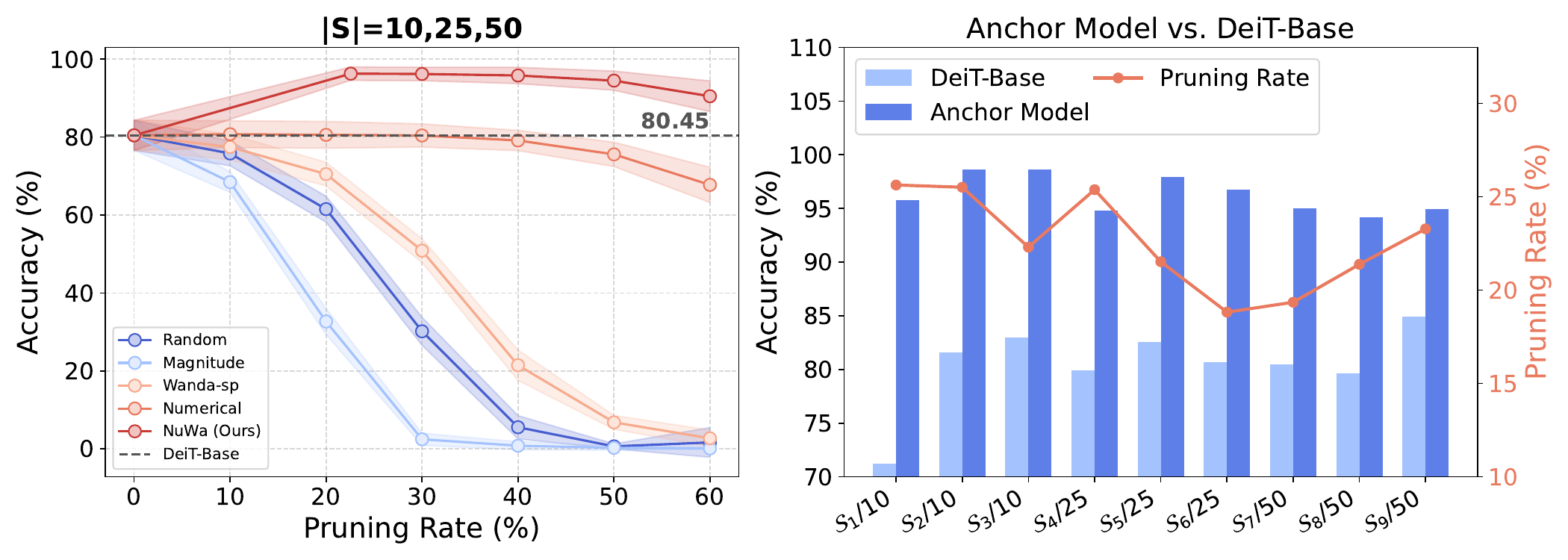}
   \caption{Comparison between NuWa and training-free pruning baselines across sub-tasks of different scales. The lightweight models derived by NuWa outperform the base ViT on target classes, which existing methods fail to achieve.}
   \label{fig:training_free_comparison}
\end{figure}

\section{Experiments and Analysis}
\label{sec:experiment}

\textbf{Models, Datasets, and Sub-Tasks}. Following prior studies~\cite{yu2023x,sun2025mdp}, we evaluate NuWa with DeiT-B/S/T and ViT-L/16~\cite{touvron2021training-icml} on ImageNet-1K~\cite{russakovsky2015imagenet}, CIFAR-100, and CIFAR-10~\cite{krizhevsky2009cifar}, Fast/Mask R-CNN (Swin-T)~\cite{liu2021swin} on COCO2017~\cite{lin2014microsoft}. Each sub-task $\mathcal{S}_i/N$ contains $N$ randomly selected classes from the corresponding dataset.

\textbf{Implementations}. During SKP, the learning rates for the mask vectors $\mathcal{M}$ and control factors $\mathcal{B}$ are set to $0.001$ and $0.1$, respectively, and the total training steps are $10^4$ with a batch size of 1. AdamW is used as the optimizer. When applying OFP, the retained energy ratio $\rho$ is adaptively determined based on the target pruning rate (\cref{sec:implementation_details} in Suppl.), and the number of calibration samples $K$ is set to 128. GFLOPs is used to reflect device resource constraints and to compute pruning rates. All experiments are conducted on an NVIDIA RTX 4090 GPU.

\textbf{Baselines}. We first compare NuWa with representative training-free structured pruning methods, including Magnitude Pruning~\cite{han2015learning}, Wanda-sp~\cite{sun2023simple,an2024fluctuation}, and Numerical Pruning~\cite{shen2025numerical}. Then, we compare NuWa with four state-of-the-art training-dependent structured pruning methods, including X-Pruner~\citep{yu2023x}, DC-ViT~\citep{zhang2024dense}, RECAP~\cite{ilhan2024resource}, and MDP~\citep{sun2025mdp}. The Random method serves as a lower-bound baseline for evaluation. See \cref{sec:baseline} in the Suppl. for more details.

\subsection{Main Results}
\label{sec:results}

\textbf{Comparison with Training-Free Baselines}. We randomly select nine sub-tasks ($\mathcal{S}_1$/10–$\mathcal{S}_9$/50) from ImageNet and compare the accuracy of edge ViTs derived from DeiT-Base by NuWa and training-free baselines under different pruning rates. As shown in \cref{fig:training_free_comparison}, edge ViTs derived by baselines fail to surpass the base ViT because they retain class-detrimental knowledge. In contrast, NuWa eliminates such knowledge through SKP, producing an anchor model that achieves an average pruning rate of 22.61\% while improving the average class-specific accuracy by 15.84\%. Compared with the base ViT and the best-performing Numerical Pruning, NuWa achieves an accuracy increase of up to 20.60\% and 29.00\%, respectively, at a pruning rate of 0.60. These results demonstrate that NuWa effectively takes advantage of the “free lunch” within ViTs, deriving smaller yet more accurate class-specific models without retraining.


\begin{table*}
\centering
\caption{Comparison in derivation efficiency between NuWa and training-dependent baselines across different sub-tasks scales. $\mathcal{P}$ and $\mathcal{T}$ denote pruning and retraining costs. Accuracy is averaged over $\mathcal{S}_1$-$\mathcal{S}_9$ in \cref{tab:training_dependent_comparison}, and cost is based on AWS EC2 g5.48xlarge pricing.}
\vskip -0.1in
\label{tab:derivation_efficiency}
\begin{center}
\begin{scriptsize}
\renewcommand{\arraystretch}{1.1}
\begin{tabular}{l|ccc|c|ccc|c|c}
\toprule
\multirow{2}{*}{\textbf{Methods}} & 
\multicolumn{4}{c|}{\raisebox{0.5\totalheight}{\textbf{Overhead (GPU Hours)}}} &
\multicolumn{4}{c|}{\raisebox{0.5\totalheight}{\textbf{AWS Cost ($N=50$,~$M=10$)}}} & 
\multirow{2}{*}{\textbf{Acc. (\%)}}\\
\cline{2-9}
& $|\mathcal{S}|$=10 & $|\mathcal{S}|$=25 & $|\mathcal{S}|$=50 & Formula & $|\mathcal{S}|$=10 & $|\mathcal{S}|$=25 & $|\mathcal{S}|$=50 & \textit{Avg.}\\
\hline\hline
\textbf{X-Pruner ($\mathcal{P+T}$)} & 0.99 & 2.50 & 5.01 & $MN\times(\mathcal{P}+\mathcal{T})$ & \$8063 & \$20360 & \$40801 & \$23074 & \textbf{93.77}\\
\rowcolor[gray]{0.95}
\textbf{DC-ViT ($\mathcal{P+T}$)} & 2.08 & 2.02 & 2.13 & $MN\times(\mathcal{P}+\mathcal{T})$ & \$16939 & \$17265 & \$17347 & \$17184 & 80.68\\
\textbf{RECAP ($\mathcal{P+T}$)} & 0.89 & 2.25 & 4.51 & $MN\times(\mathcal{P}+\mathcal{T})$ & \$7248 & \$18324 & \$36729 & \$20767 & 93.13\\
\rowcolor[gray]{0.95}
\textbf{MDP ($\mathcal{P+T}$)} & 2.26 & 5.61 & 11.27 & $MN\times(\mathcal{P}+\mathcal{T})$ & \$18405 & \$45688 & \$91783 & \$51959 & 92.95\\
\hline
\textbf{\textit{NuWa} ($\mathcal{P}_1$:~\cref{eq:prune_mha})} & 0.01 & 0.01 & 0.01 & 
\multirow{4}{*}{
  $\begin{array}{c}
  \mathcal{P}_1 + N\times\mathcal{P}_2 \\
  +\, MN\times\mathcal{P}_3
  \end{array}$
} & \$0.16 & \$0.16 & \$0.16 & \$0.16 & 
\multirow{4}{*}{
  $\begin{array}{c}
  93.16 \\
  \textcolor{downblue}{(\downarrow 0.61)}
  \end{array}$
}
\\
\cellcolor[gray]{0.95}\textbf{\textit{NuWa} ($\mathcal{P}_2$:~\cref{eq:get_anchor_model}, $\mathcal{H}^{(l)}$)} & \cellcolor[gray]{0.95}0.06 & \cellcolor[gray]{0.95}0.07 & \cellcolor[gray]{0.95}0.08 & & \cellcolor[gray]{0.95}\$48.86 & \cellcolor[gray]{0.95}\$57.01 & \cellcolor[gray]{0.95}\$65.15 & \cellcolor[gray]{0.95}\$57.01 &\\
\textbf{\textit{NuWa} ($\mathcal{P}_3$:~\cref{eq:prune_mlp})} & 2.28e-4 & 2.39e-4 & 2.50e-4 & & \$1.86 & \$1.95 & \$2.04 & \$1.95 &\\
\cellcolor[gray]{0.95}\textbf{\textit{NuWa} ($\mathcal{P}_1+\mathcal{P}_2+\mathcal{P}_3$)} & \cellcolor[gray]{0.95} 0.07 & \cellcolor[gray]{0.95} 0.08 & \cellcolor[gray]{0.95} 0.09 & & \cellcolor[gray]{0.95}\$50.88 & \cellcolor[gray]{0.95}\$59.12 & \cellcolor[gray]{0.95}\$67.35 & \cellcolor[gray]{0.95}\textbf{\$59.12} &\\
\hline\bottomrule
\end{tabular}
\end{scriptsize}
\end{center}
\vskip -0.3in
\end{table*}

\textbf{Comparison with Training-dependent Baselines}. \cref{tab:training_dependent_comparison} reports the accuracy of the edge ViTs derived from DeiT-Base by NuWa and training-dependent baselines. Compared with the original DeiT-Base, NuWa achieves average class-specific accuracy improvements of 15.37\% and 10.04\% at pruning rates of 0.40 and 0.60, respectively, without any retraining. When compared with the best performing baselines, i.e., X-Pruner, RECAP, and MDP, the models derived by NuWa reach 99.35\%, 100.03\%, and 100.23\% of their average accuracy, respectively. With light fine-tuning (10 epochs as in RECAP), NuWa further surpasses them by 2.00\%, 2.64\%, and 2.82\% on average. For efficiency, we simulate a deployment scenario with $N$ sub-tasks and $M$ heterogeneous edge devices per sub-task, requiring $MN$ models. As shown in \cref{tab:derivation_efficiency}, NuWa achieves an average speedup of 33.69× over X-Pruner, the best-performing baseline, for a single model with only 0.61\% accuracy loss. Benefiting from reusable pruning results and calibration features that can be shared across different sub-tasks and devices (\cref{sec:analysis}), NuWa significantly reduces the derivation overhead in large-scale deployments. When $N$=50 and $M$=10, NuWa reduces the derivation time and cost by up to 99.83\% compared with X-Pruner, and by 99.70\%, 99.82\%, and 99.93\% compared with other baselines.

\begin{figure}[t]
   \centering
   \includegraphics[width=1.0\linewidth]{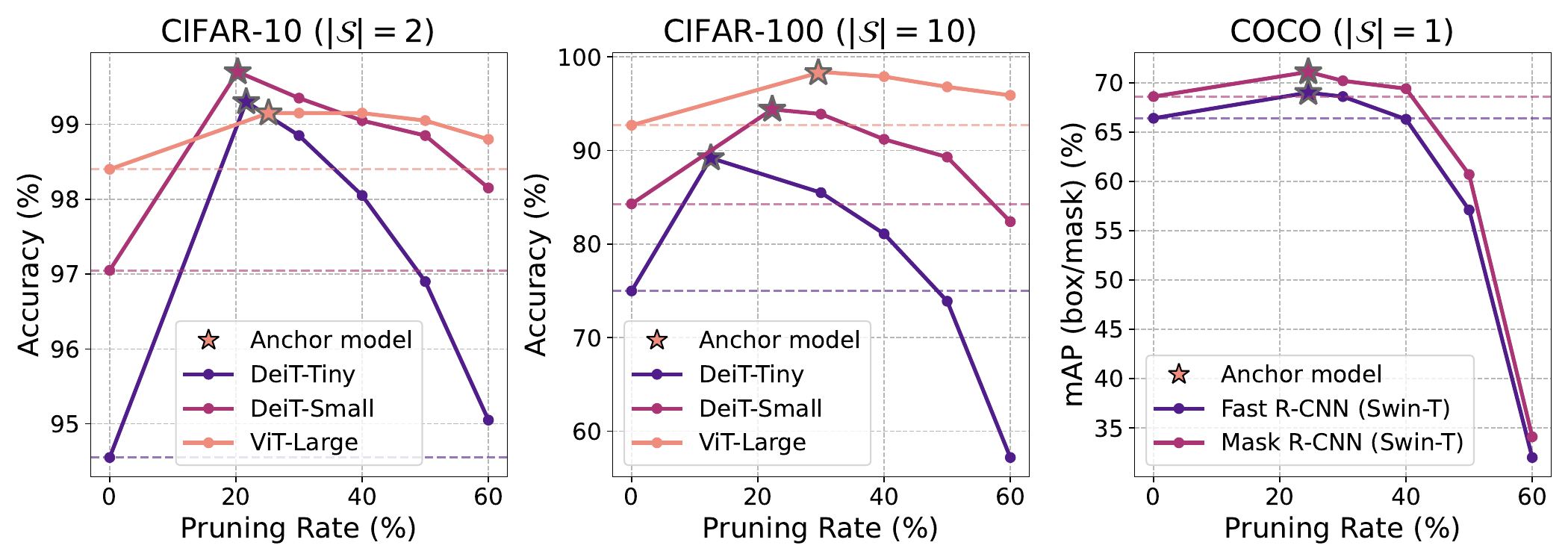}
   \caption{Performance of NuWa-derived edge ViTs across different base ViTs and datasets, with randomly selected classes in $\mathcal{S}$.}
   \label{fig:generality}
\end{figure}

\textbf{Generality Across Models and Datasets}. To evaluate the generality of NuWa, we extend the experiments beyond DeiT-Base on ImageNet to include DeiT-Small/Tiny and ViT-Large on CIFAR-10 and CIFAR-100, and Fast/Mask R-CNN (Swin-T) on COCO. As shown in \cref{fig:generality}, for recognition tasks, the edge ViTs derived from DeiT-Tiny, DeiT-Small, and ViT-Large consistently outperform the corresponding base ViTs on target classes when the pruning rate is below 0.40, 0.50, and 0.60, respectively. We observe that as the model size increases, the pruning rate of the anchor model also increases, indicating a higher proportion of class-detrimental weights. For more complex detection and segmentation tasks, NuWa can also derive models that outperform the base ViT in mAP on target classes when the backbone pruning rate is below 0.40.

\begin{table*}[t]
\centering
\caption{Comparison in computational efficiency between DeiT-Base and edge ViTs derived by NuWa under different $\alpha$.}
\label{tab:speedup}
\begin{scriptsize}
\renewcommand{\arraystretch}{1.1}
\begin{tabular}{l||cc|cc|cc|c|c}
\toprule
\multirow{2}{*}{\textbf{Methods}} &
\multicolumn{2}{c|}{\raisebox{0.5\totalheight}{\textbf{Latency (ms)}}} & 
\multicolumn{2}{c|}{\raisebox{0.5\totalheight}{\textbf{Throughput (image/s)}}} &
\multicolumn{2}{c|}{\raisebox{0.5\totalheight}{\textbf{Memory (GB)}}} &
\multicolumn{1}{c|}{\raisebox{0.5\totalheight}{\textbf{\#Param}}} & 
\multicolumn{1}{c}{\raisebox{0.5\totalheight}{\textbf{FLOPs}}}\\
\cline{2-7}
& Orin NX & RTX 4090 & Orin NX & RTX 4090 & Orin NX & RTX 4090 & (M) & (G)\\
\midrule
\textbf{DeiT-Base} & 45.45 & 274.27 & 22.00 & 933.39 & 0.34 & 2.21 & 86.57 & 17.57\\
\midrule
\rowcolor[gray]{0.95}
\textbf{NuWa ($\mathcal{V}_E=\mathcal{V}_A$)} & 34.84 (1.31$\times$) & 211.86 (1.30$\times$) & 28.70 & 1208.34 & 0.25\textsubscript{\textcolor{downblue}{$\downarrow$26.47\%}} & 1.93\textsubscript{\textcolor{downblue}{$\downarrow$12.67\%}} & 66.90\textsubscript{\textcolor{downblue}{$\downarrow$22.72\%}} & 13.60\textsubscript{\textcolor{downblue}{$\downarrow$22.61\%}}\\
\textbf{NuWa ($\alpha(\mathcal{V}_E)=0.40$)} & 29.70 (1.53$\times$) & 186.01 (1.47$\times$) & 33.67 & 1376.27 & 0.21\textsubscript{\textcolor{downblue}{$\downarrow$38.24\%}} & 1.57\textsubscript{\textcolor{downblue}{$\downarrow$28.96\%}} & 51.75\textsubscript{\textcolor{downblue}{$\downarrow$40.14\%}} & 10.53\textsubscript{\textcolor{downblue}{$\downarrow$40.07\%}}\\
\rowcolor[gray]{0.95}
\textbf{NuWa ($\alpha(\mathcal{V}_E)=0.60$)} & 22.01 (2.07$\times$) & 142.52 (1.92$\times$) & 45.43 & 1796.23 & 0.14\textsubscript{\textcolor{downblue}{$\downarrow$58.82\%}} & 1.29\textsubscript{\textcolor{downblue}{$\downarrow$41.63\%}} & 34.48\textsubscript{\textcolor{downblue}{$\downarrow$60.17\%}} & 7.00\textsubscript{\textcolor{downblue}{$\downarrow$60.16\%}}\\
\bottomrule
\end{tabular}
\end{scriptsize}
\vskip -0.15in
\end{table*}

\textbf{Inference Speedups}. We evaluate the speedups of edge ViTs derived by NuWa on two representative devices, i.e., Jetson Orin NX and NVIDIA RTX 4090. Considering the difference in real-world usage, we use a batch size of 1 for Orin NX to simulate sequential real-time requests and a batch size of 256 for RTX 4090 to simulate server-side batch processing. The results are averaged over the nine models from \cref{tab:training_dependent_comparison}. As shown in \cref{tab:speedup}, compared with the base ViT, NuWa achieves a speedup of 1.31$\times$-2.07$\times$ on Orin NX and 1.30$\times$-1.92$\times$ on RTX 4090, while reducing memory consumption by 26.47\%–58.82\% and 12.67\%–41.63\%, respectively.

\subsection{Analysis}
\label{sec:analysis}

\begin{figure}[t]
   \centering
   \includegraphics[width=0.95\linewidth]{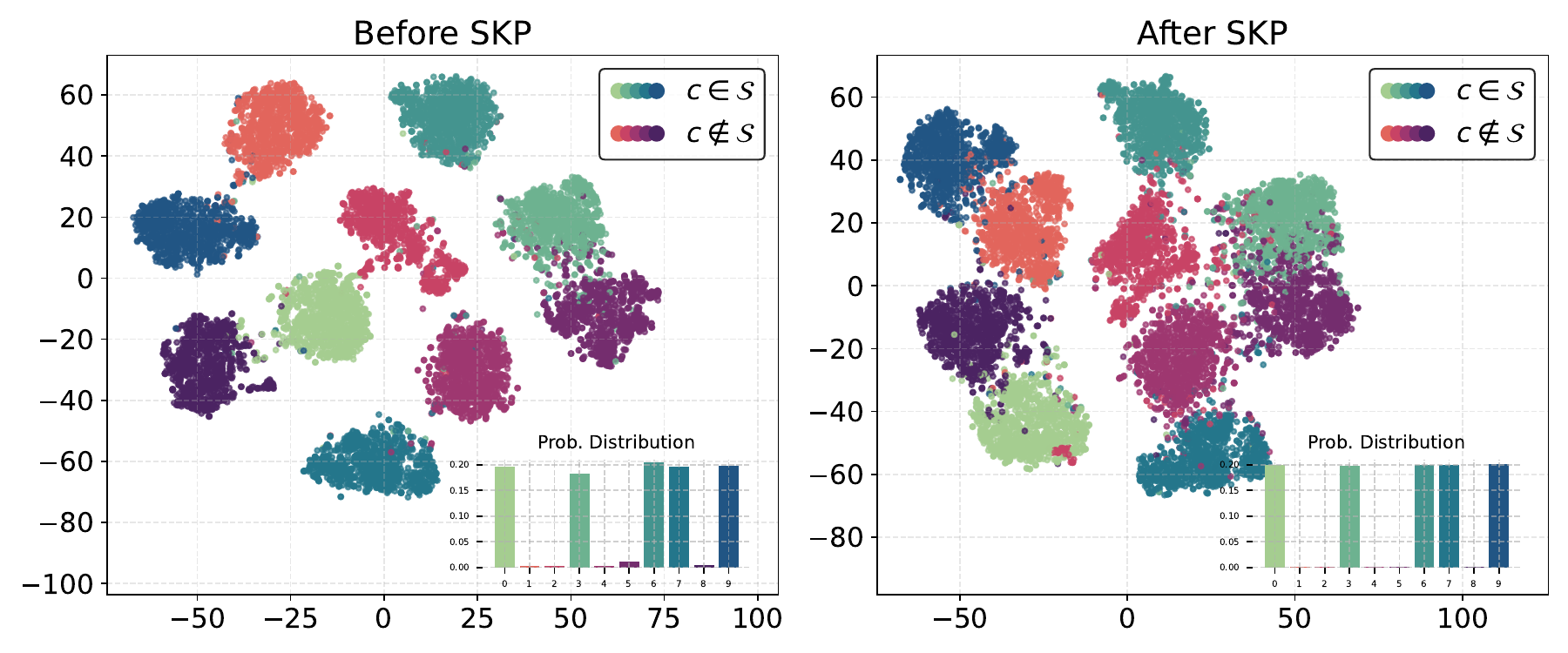}
   \vskip -0.05in
   \caption{Comparison in feature and probability distributions before and after SKP. The feature distributions are visualized by applying t-SNE to the CLS tokens from the last block, while the bar charts show the average output probability over $\mathcal{D}_{\mathcal{S}/5}$.}
   \label{fig:skp_analysis}
   \vskip -0.15in
\end{figure}

\textbf{Interpretability of SKP}. As shown in \cref{fig:training_free_comparison}, NuWa can use SKP to prune the model while improving its performance on target classes, which seems counterintuitive. To interpret this, we construct a sub-task $\mathcal{S}$ with five random classes from CIFAR-10 and apply SKP to DeiT-B fine-tuned on CIFAR-10. \cref{fig:skp_analysis} shows that SKP slightly blurs class boundaries, particularly among non-target classes in $\mathcal{S}^c$ ($\mathcal{S}^c=\mathcal{Y}\setminus \mathcal{S}$), but effectively suppresses the output probabilities of classes in $\mathcal{S}^c$. This indicates that SKP enhances the model's focus on target classes by reducing misclassification into $\mathcal{S}^c$. In practice, we further reinforce this effect by assigning large negative bias values (e.g., -100) to non-target classes in $\mathcal{V}_A$’s classifier. This fundamentally prevents $\mathcal{V}_A$ from misclassifying inputs into classes in $\mathcal{S}^c$.

\begin{figure}[t]
   \centering
   \includegraphics[width=0.95\linewidth]{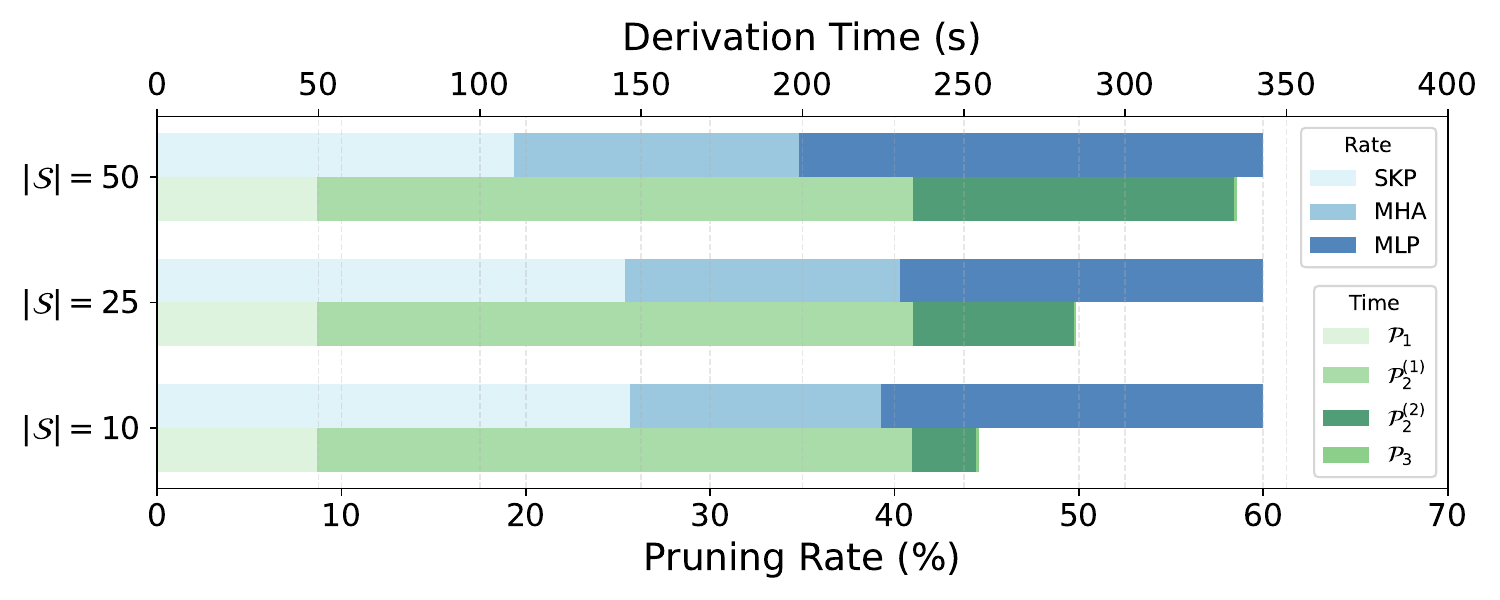}
   \vskip -0.05in
   \caption{Breakdown of pruning rates and derivation times across sub-tasks of different scales when the overall pruning rate is 0.60.}
   \label{fig:derivation_analysis}
\end{figure}

\textbf{Breakdown of Derivation}. \cref{fig:derivation_analysis} demonstrates the contributions of different modules to the overall pruning rate and the time breakdown during NuWa’s derivation process. For the pruning rate, SKP contributes an average pruning rate of 22.61\%. When the overall pruning rate reaches 0.60, OFP contributes 14.67\% and 22.72\% pruning rates by further prunes MHA and MLP modules, respectively. Most of the pruning occurs in MLPs, which account for about two-thirds of the total parameters in ViTs. The total derivation time consists of three parts, i.e., $\mathcal{P}_1$, $\mathcal{P}_2$, and $\mathcal{P}_3$. $\mathcal{P}_1$ corresponds to the SVD-based pruning of $W_{QK}^{\smash{(l)}}$ and $W_{VO}^{\smash{(l)}}$, which is data-free and can be reused across different sub-tasks and pruning rates. $\mathcal{P}_2$ includes SKP ($\mathcal{P}_2^{\smash{(1)}}$) and one-epoch forward propagation on $\mathcal{D_S}$ for computing activations $a^{\smash{(l)}}$ and features $\mathcal{H}^{\smash{(l)}}$ ($\mathcal{P}_2^{\smash{(2)}}$). This stage dominates the total computation, but can be reused across different pruning rates. $\mathcal{P}_3$ prunes MLP based on \cref{eq:prune_mlp}. Although its results are not reusable, its computational cost is negligible.

\begin{table}[t]
    \caption{Ablation study of pruning techniques and settings used by NuWa on DeiT-Base with $\alpha = 0.6$.}
    \vskip -0.1in
    \label{tab:ablation_study}
    \begin{center}
    \begin{scriptsize}
    \renewcommand{\arraystretch}{1.1}
    \begin{tabular}{l||ccc|c}
    \toprule
    \textbf{Setting} & $\mathcal{S}_4$/25 & $\mathcal{S}_5$/25 & $\mathcal{S}_6$/25 & \textbf{\textit{Avg}}\\
    \hline\hline
    \textbf{NuWa} & 89.84 & 92.00 & 92.48 & \textbf{91.44}\\
    \midrule
    \rowcolor[gray]{0.95}
    \textbf{w/o SKP} & 69.04 & 76.72 & 72.32 & 72.69\textsubscript{\textcolor{downblue}{$\downarrow$18.75}}\\
    \textbf{w/o MHA pruning} & 74.48 & 75.60 & 78.08 & 76.05\textsubscript{\textcolor{downblue}{$\downarrow$15.39}}\\
    \rowcolor[gray]{0.95}
    \textbf{w/o MLP pruning} & 4.72 & 7.76 & 4.00 & 5.49\textsubscript{\textcolor{downblue}{$\downarrow$85.95}}\\
    \textbf{w/o Activation} & 84.64 & 87.84 & 86.88 & 86.45\textsubscript{\textcolor{downblue}{$\downarrow$4.99}}\\
    \rowcolor[gray]{0.95}
    \textbf{w/o Optimization} & 75.20 & 76.08 & 78.80 & 76.69\textsubscript{\textcolor{downblue}{$\downarrow$14.75}}\\
    \bottomrule
    \end{tabular}
    \end{scriptsize}
    \end{center}
\vskip -0.15in
\end{table}

\textbf{Ablation Study}. We validate the necessity of each design in NuWa on $\mathcal{S}_4/25$-$\mathcal{S}_6/25$ in \cref{tab:training_dependent_comparison}. The SKP is essential for enabling NuWa to derive small models that outperform the base ViT. As shown in \cref{tab:ablation_study}, removing SKP prevents the model from focusing on target classes, causing an average accuracy drop of 18.75\%. When OFP skips pruning the MHA or MLP modules, accuracy decreases by 15.39\% and 85.95\%, respectively. This highlights the necessity of balancing their pruning intensities through $\rho$. Randomly determining $\mathcal{I}_r^{(l)}$ instead of using $a^{(l)}$ lowers accuracy by 4.99\%. Replacing the closed-form solution in \cref{eq:mlp_pruning_problem} with direct removal of low-activation rows in $W_2^{\smash{(l)}}$ reduces accuracy by 14.75\%. This demonstrates the advantage of NuWa in preserving class-specific knowledge.

\begin{figure}[t]
   \centering
   \includegraphics[width=0.95\linewidth]{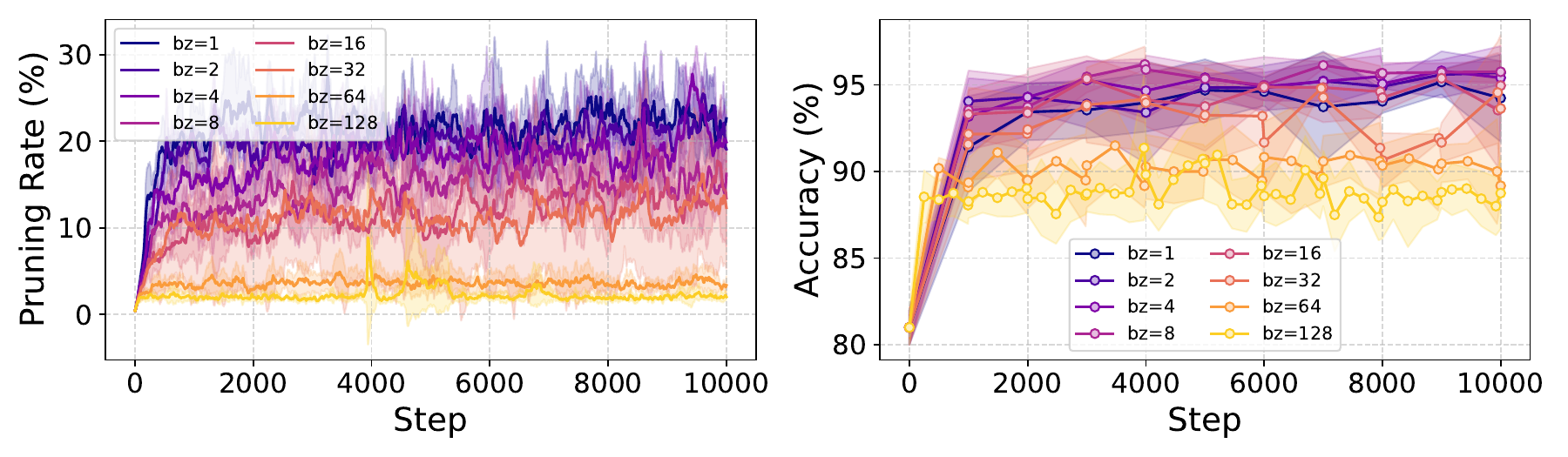}
   \caption{Pruning rate and accuracy of the anchor model $\mathcal{V}_A$ during SKP under different batch sizes.}
   \label{fig:batch_size}
   \vskip -0.15in
\end{figure}

\begin{figure}[t]
    \centering
    \begin{minipage}{0.45\linewidth}  
        \centering
        \includegraphics[width=0.95\linewidth]{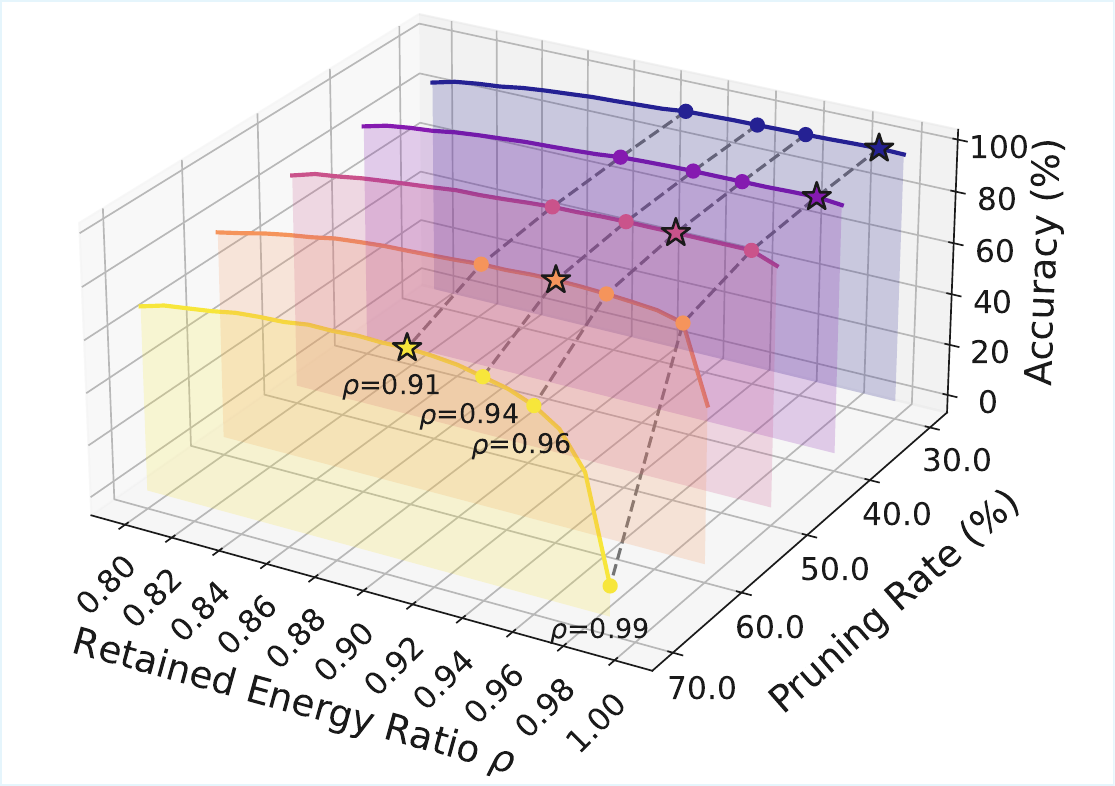}
        \caption{Effect of the retained energy ratio $\rho$.}
        \label{fig:rho_analysis}
    \end{minipage}
    \hspace{0.02\linewidth}
    \begin{minipage}{0.48\linewidth}
        \centering
        \includegraphics[width=0.95\linewidth]{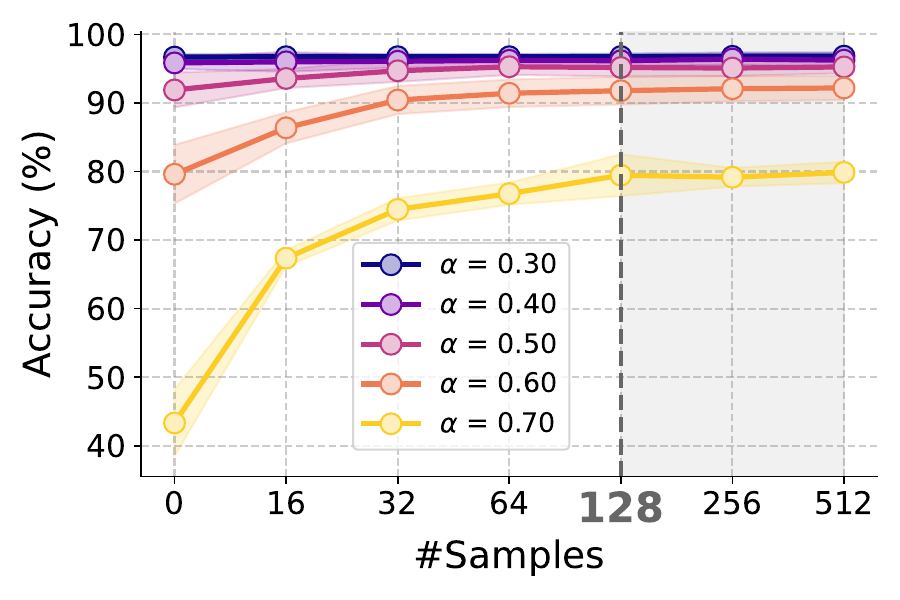}
        \caption{Effect of the number of calibration samples $K$.}
        \label{fig:sample_number}
    \end{minipage}
\end{figure}

\textbf{Hyperparameter Analysis}. We analyze the effect of key hyperparameters on NuWa’s performance. Unless otherwise specified, all experiments are conducted on DeiT-Base with a pruning rate of 0.60, with the sub-tasks $\mathcal{S}_4$-$\mathcal{S}_6$ in \cref{tab:training_dependent_comparison}. As shown in \cref{fig:batch_size}, a smaller batch size during SKP leads to a higher pruning rate of $\mathcal{V}_A$, enabling $\mathcal{V}_B$ to explore the pruning space more thoroughly. Therefore, we set the batch size to 1, which also improves derivation efficiency. \cref{fig:rho_analysis} illustrates the effect of different retained energy ratios $\rho$ on the accuracy of NuWa-derived models across different pruning rates $\alpha$. As $\alpha$ increases, the optimal $\rho$ (marked with asterisks) gradually decreases. This indicates that a larger proportion of pruning should be allocated to the MHA modules. Since $\mathcal{P}_3$ is negligible, NuWa allows efficient search for a proper $\rho$ value in practice to determine the optimal pruning configuration. \cref{fig:sample_number} shows that activation features $\mathcal{H}^{(l)}$ computed from 128 randomly sampled (\cref{sec:design_justification} in Suppl.) images are sufficient for NuWa to obtain optimal closed-form solutions of $W_2^{\smash{(l)\prime}}$.

%% file: sec/5_conclusion.tex
\section{Conclusion}
\label{sec:conclusion}

This paper presented NuWa, a cost-efficient model derivation method that can derive lightweight class-specific ViTs from pre-trained base ViTs for edge devices. Motivated by the insight that removing certain weights elevates target-class performance, NuWa introduced Self-Knowledge Purification to identify and prune such class-detrimental weights, producing a refined anchor model. Then, NuWa solved optimization problems on the anchor model to directly obtain the closed-form solutions of pruned weights efficiently. Without retraining, NuWa achieved comparable accuracy to computation-intensive structured pruning baselines with the same pruning rate. It significantly reduced the cost of large-scale class-specific model deployment. Extensive experiments with six models on four datasets demonstrated its effectiveness, efficiency, and generality.

%% file: sec/6_acknowledgment.tex
\section*{Acknowledgments}
\label{sec:acknowledgments}

This research was supported by the National Key R\&D Program of China under Grant No.~2023YFB4502400.

%% file: sec/X_suppl.tex
\clearpage
\setcounter{page}{1}
\maketitlesupplementary

\section{Calculation of FLOPs}
\label{sec:gflops}

We adopt GFLOPs, which is widely used to measure the computational cost of model inference, to reflect the resource constraints of edge devices~\cite{abadi2016tensorflow}. FLOPs represent the total number of floating-point operations required for a model to perform inference on a single input. For a Vision Transformer (ViT), let the embedding dimension be $d$; the query–key dimension, value–output dimension, and number of heads in the MHA module of the $l$-th block be $q_l$, $v_l$, and $H_l$, respectively; and let the intermediate dimension of the MLP module in the same block be $e_l$. Given an input $\mathbf{X}\in\mathbb{R}^{N\times d}$ consisting of $N$ patch tokens, the total FLOPs of a ViT with $L$ blocks can be formulated as:
\begin{equation}
\label{eq:flops}
\text{FLOPs}=(2Nd+N^2)\sum_{l=1}^LH_l(q_l+v_l)+2Nd\sum_{l=1}^Le_l
\end{equation}

\section{Limitations of Compression Methods}
\label{sec:limitations_of_compression_methods}

\begin{table}[t]
\caption{Comparison in ViT throughput (images/s) on Jetson Orin NX before and after INT8 quantization and magnitude pruning.}
\label{tab:quant_unstructured_latency}
\begin{center}
\begin{scriptsize}
\renewcommand{\arraystretch}{1.1}
\begin{tabular}{l||c|c|c}
\toprule
\multirow{2}{*}{\textbf{Methods}} &
\multicolumn{1}{c|}{\raisebox{0.5\totalheight}{\textbf{Base ViT}}} & 
\multicolumn{1}{c|}{\raisebox{0.5\totalheight}{\textbf{INT8}}} &
\multicolumn{1}{c}{\raisebox{0.5\totalheight}{\textbf{Magnitude}}}\\
\cline{2-4}
& (GPU) & (CPU) & (GPU)\\
\midrule\hline
\textbf{DeiT-Base} & 22.00 & 0.29\textsubscript{\textcolor{downblue}{$\downarrow$98.68\%}} & 21.99\textsubscript{\textcolor{downblue}{$\downarrow$0.05\%}}\\
\textbf{DeiT-Small} & 64.16 & 0.67\textsubscript{\textcolor{downblue}{$\downarrow$98.96\%}} & 63.28\textsubscript{\textcolor{downblue}{$\downarrow$1.37\%}}\\
\textbf{DeiT-Tiny} & 68.87 & 1.50\textsubscript{\textcolor{downblue}{$\downarrow$98.48\%}} & 67.21\textsubscript{\textcolor{downblue}{$\downarrow$2.41\%}}\\
\bottomrule
\end{tabular}
\end{scriptsize}
\end{center}
\end{table}

\textbf{Low-bit Quantization and Unstructured Pruning}. As discussed in \cref{sec:related_works}, quantization and unstructured pruning rely heavily on specific hardware and software architectures to achieve real acceleration. To illustrate this, we apply INT8 quantization~\citep{xi2024jetfire} and Magnitude Pruning (with a pruning ratio of 0.50)~\citep{han2015learning}, two representative techniques of quantization and unstructured pruning, to compress DeiT-Base, DeiT-Small, and DeiT-Tiny. Then, we measure the inference throughput of compressed ViTs on a widely used edge device, Jetson Orin NX, with a batch size of 1. As shown in \cref{tab:quant_unstructured_latency}, due to the lack of TensorRT support, the INT8-quantized model can only perform inference on the CPU, resulting in a 98.71\% decrease in average throughput. Similarly, the model obtained by unstructured pruning shows almost no acceleration, as current frameworks lack libraries optimized for sparse matrix operations. These results indicate that both methods suffer from poor adaptability when deployed on edge devices with highly heterogeneous frameworks and drivers.

\begin{figure}[t]
   \centering
   \includegraphics[width=0.9\linewidth]{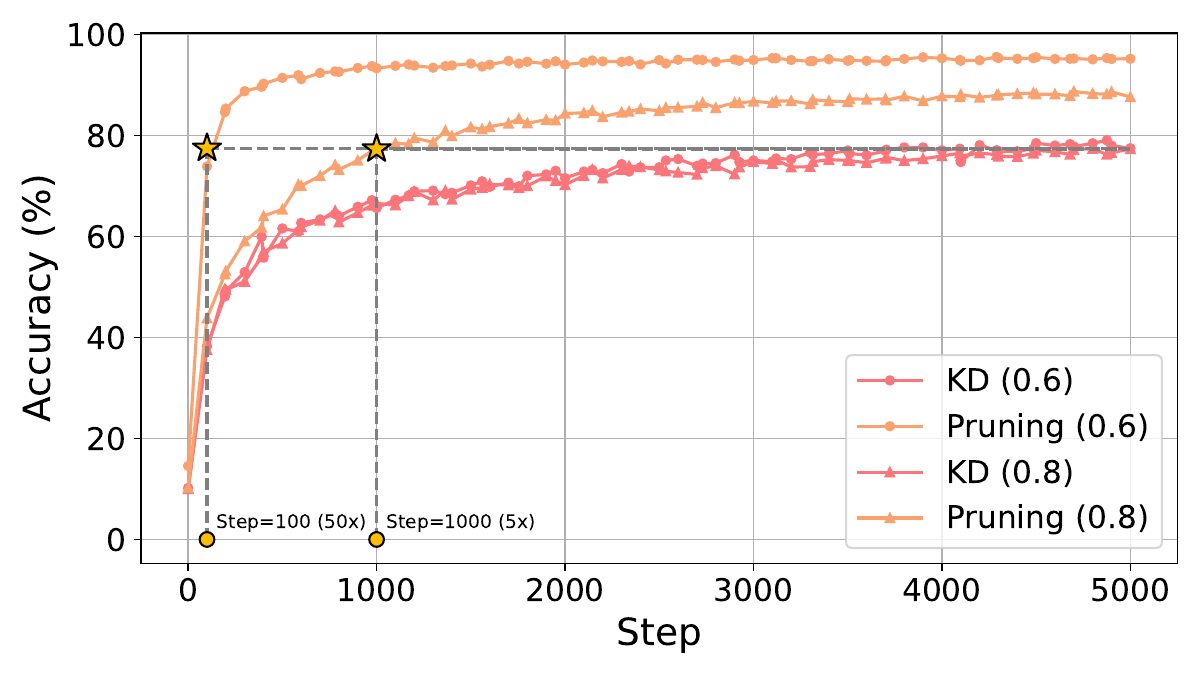}
   \caption{Comparison of convergence speed between logit-based knowledge distillation (KD) and random pruning (Pruning) across different pruning rates with DeiT-Base on CIFAR-10.}
   \label{fig:kd_vs_pruning}
\end{figure}

\textbf{Knowledge Distillation}. Knowledge distillation (KD) focuses on transferring the teacher model’s knowledge to the student model in the feature space and does not provide a good initialization in the parameter space for arbitrary student architectures. As a result, the student model often needs to be trained from scratch, leading to slow convergence and high computational cost. To illustrate this, we compare the convergence of logit-based KD~\citep{hinton2015distilling} and random structured pruning~\citep{gadhikar2023random} with DeiT-Base on CIFAR-10~\citep{krizhevsky2009learning}. Specifically, we use student models that share the same architecture as the pruned ones but are randomly initialized. The teacher model is a DeiT-Base pretrained and fine-tuned on CIFAR-10. As shown in \cref{fig:kd_vs_pruning}, the pruned models converge significantly faster than their KD counterparts, highlighting the importance of parameter-space knowledge transfer.

Although the above methods have clear limitations, they are orthogonal to NuWa and can be used together when supported by edge devices to achieve further compression.

\section{Importance Metrics for Pruning}
\label{sec:metric}

\textbf{Magnitude}. Magnitude pruning is a simple and widely used pruning approach in model compression~\cite{han2015learning}. The basic idea is to prune the weights with the smallest magnitudes, assuming that smaller weights contribute less to the model's output and, thus, can be removed without significantly affecting performance. In the case of pruning according to the magnitude of weights, the importance score is calculated as $I(W)=|W|=\sum_{i}|w_i|$.

\textbf{Activation}. Activation pruning measures the importance of each structure by analyzing its response to input data~\cite{chen2021chasing,sun2023simple}. The importance score is calculated as $I(W)=\sum_i a_i$. Structures with lower average activation values are considered less important for the current task. Since it relies on input data, activation pruning can capture features specific to certain classes, making it well-suited for our class-specific model derivation.

\textbf{Gradient}. During backward propagation, gradients indicate how sensitive the loss function is to changes in each weight~\cite{lee2018snip,yu2018nisp}. Weights with smaller gradients are considered less important for the model’s predictions and can be pruned with minimal impact on performance. The importance score is calculated as $I(W)=|\frac{\mathcal{L}(x)}{\partial W}|=\sum_i|\frac{\mathcal{L}(x)}{\partial w_i}|$. 

\textbf{Taylor Expansion Approximation}. Taylor pruning assumes that a smaller change in the loss value after removing a weight indicates lower importance of that weight for prediction~\cite{molchanov2019pruning,yang2023global}. However, accurately evaluating the importance of all $N$ weights would require $N$ forward propagations to compute the corresponding loss changes, which is computationally expensive. To reduce this overhead, these methods adopt the following approximation based on Taylor expansion:
\begin{equation}
\label{eq:taylor}
\begin{aligned}
I(w) &= |\mathcal{L}(x)-\mathcal{L}_{w=0}(x)|\\
     &= |\mathcal{L}(x)-\big(\mathcal{L}(x)-\frac{\partial\mathcal{L}(x)}{\partial w} w+R(w)\big)|\\
     &\overset{R(w)\approx 0}{\approx} \bigg|\frac{\partial\mathcal{L}(x)}{\partial w}w\bigg| 
\end{aligned}
\end{equation}
This approximation reduces the complexity of evaluating the importance of all weights from $O(N)$ to $O(1)$.

\section{Ablation and Design Justification}
\label{sec:design_justification}

\begin{table}[t]
    \caption{Accuracy and pruning rate (Acc./Rate) of the anchor models under different SKP application settings.}
    \label{tab:skp_modules}
    \begin{center}
    \begin{scriptsize}
    \renewcommand{\arraystretch}{1.1}
    \begin{tabular}{l||ccc|c}
    \toprule
    \textbf{Setting} & $\mathcal{S}_4$/25 & $\mathcal{S}_5$/25 & $\mathcal{S}_6$/25 & \textbf{\textit{Avg}}\\
    \hline\hline
    \rowcolor[gray]{0.95}
    \textbf{MLP Only} & 94.80/25.38 & 97.92/21.52 & 96.72/18.82 & \textbf{96.48/21.91}\\
    \textbf{MHA Only} & 81.20/7.38 & 82.80/7.13 & 80.80/8.87 & 81.60/7.79\\
    \rowcolor[gray]{0.95}
    \textbf{MHA + MLP} & 93.52/30.05 & 95.52/28.01 & 94.08/23.54 & 94.37/27.20\\
    \bottomrule
    \end{tabular}
    \end{scriptsize}
    \end{center}
\end{table}

\textbf{Application of SKP}. As discussed in \cref{sec:self_knowledge_purification}, NuWa does not apply SKP to the MHA modules. We conduct experiments to justify this design. Specifically, we apply SKP to different modules of DeiT-Base for the three sub-tasks $\mathcal{S}_4$/25-$\mathcal{S}_6$/25 in Tab.~1 under three settings, i.e., applying SKP only to MLPs (MLP only), only to MHAs (MHA only), and to both (MHA+MLP). As shown in \cref{tab:skp_modules}, when SKP is applied to the MHA modules to prune attention heads, the accuracy of the resulting anchor model decreases, regardless of whether SKP is also applied to the MLP modules. This is because SKP fails to control the pruning rate of the base ViT, leading to the loss of class-relevant knowledge. These results indicate that class-detrimental knowledge mainly resides in the MLP modules, while applying SKP to the MHA interferes with the model’s ability to locate and filter such knowledge. This observation is consistent with our SVD-based pruning of the MHA module, since SVD is data-free and suggests that the MHA stores class-agnostic general knowledge.

\begin{figure}[t]
   \centering
   \includegraphics[width=0.9\linewidth]{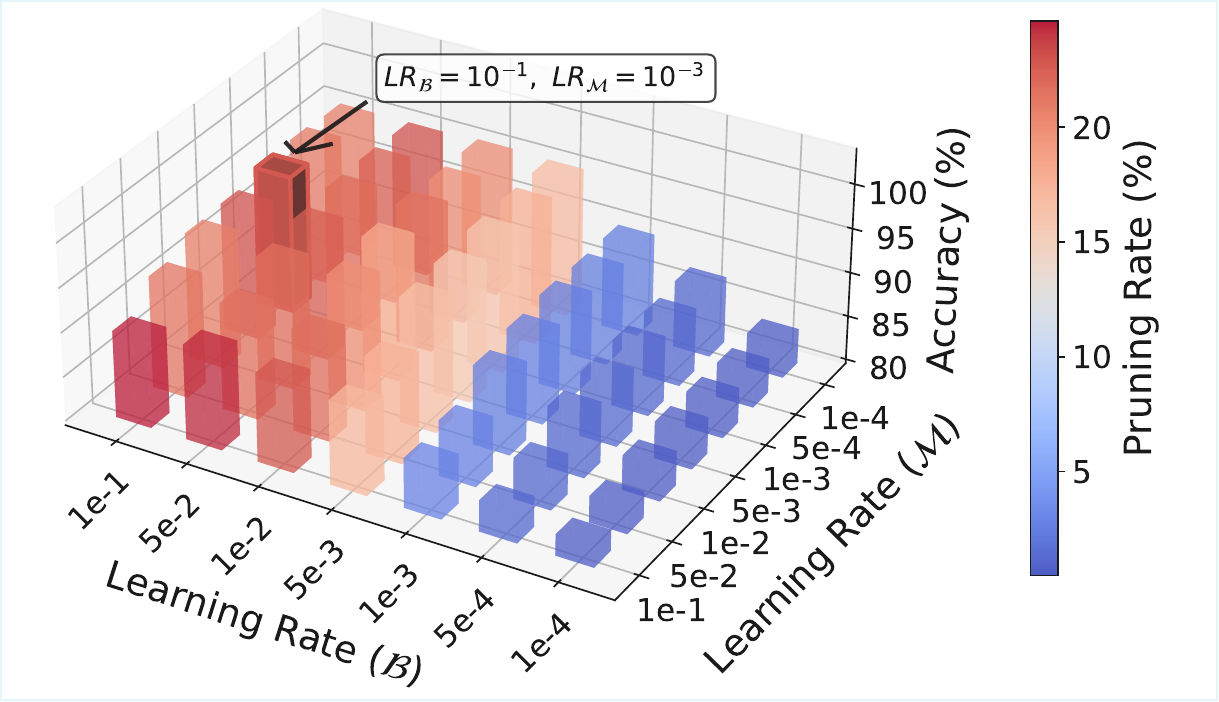}
   \caption{Effect of learning rates for $\mathcal{M}$ and $\mathcal{B}$ on the accuracy and pruning rate of the anchor model.}
   \label{fig:mask_beta_lr}
\end{figure}

\textbf{Learning Rate of $\mathcal{M}$ and $\mathcal{B}$}. During SKP, the learning rates of the mask vectors $\mathcal{M}$ and control factors $\mathcal{B}$ significantly influence how effectively class-detrimental knowledge is filtered. We evaluate different learning rate combinations on DeiT-Base for sub-tasks $\mathcal{S}_4$–$\mathcal{S}_6$, and report the average accuracy and pruning rate of the resulting anchor models $\mathcal{V}_A$. As shown in \cref{fig:mask_beta_lr}, a larger learning rate for $\mathcal{B}$ leads to more thorough filtering of class-detrimental knowledge, while a slightly smaller learning rate for $\mathcal{M}$ yields more precise localization of such detrimental weights. We therefore set $\text{LR}_{\mathcal{M}} = 0.001$ and $\text{LR}_{\mathcal{B}} = 0.1$, which achieve the highest $\mathcal{V}_A$ with a balanced pruning rate.

\begin{table}[t]
    \caption{Comparison of different strategies for determining $e_l$ during Optimized-based Fast Pruning (OFP).}
    \label{tab:target_architecture}
    \begin{center}
    \begin{scriptsize}
    \renewcommand{\arraystretch}{1.1}
    \begin{tabular}{l||ccc|c}
    \toprule
    \textbf{Setting} & $\mathcal{S}_4$/25 & $\mathcal{S}_5$/25 & $\mathcal{S}_6$/25 & \textbf{\textit{Avg}}\\
    \hline\hline
    \multicolumn{5}{l}{\textit{\textcolor{gray!70}{\textbf{Pruning Rate = 0.40}}}}\\
    \rowcolor[gray]{0.95}
    \textbf{Uniform} & 94.40 & 97.04 & 96.72 & \textbf{96.05}\\
    \textbf{Proportion} & 95.57 & 96.85 & 95.57 & 96.00\\
    \rowcolor[gray]{0.95}
    \textbf{Adaptive} & 95.49 & 96.53 & 95.81 & 95.94\\
    \hline\hline
    \multicolumn{5}{l}{\textit{\textcolor{gray!70}{\textbf{Pruning Rate = 0.60}}}}\\
    \rowcolor[gray]{0.95}
    \textbf{Uniform} & 89.84 & 92.00 & 92.48 & \textbf{91.44}\\
    \textbf{Proportion} & 88.72 & 92.28 & 91.65 & 90.88\\
    \rowcolor[gray]{0.95}
    \textbf{Adaptive} & 88.88 & 92.20 & 91.57 & 90.88\\
    \bottomrule
    \end{tabular}
    \end{scriptsize}
    \end{center}
\end{table}

\textbf{Target Architecture}. Before applying OFP, NuWa needs to determine the target model architecture, i.e., the sizes of $q_l$, $v_l$, and $e_l$. For the MHA modules, For the MHA modules, NuWa controls $q_l$ and $v_l$ through the retained energy rate $\rho$. For the MLP modules, three strategies are considered for determining $e_l$:
\begin{itemize}[leftmargin=0.2in, topsep=0.01in, itemsep=0.0in]
    \item \textit{Uniform}: pruning $e_l$ across all blocks to similar sizes to avoid excessive compression in specific MLP modules.
    \item \textit{Proportion}: allocating the total number of pruned neurons $e_\text{prune}$ to each block proportionally based on the original $e_l$ of the anchor model.
    \item \textit{Adaptive}: globally pruning the $e_\text{prune}$ neurons with the smallest normalized activation values across all blocks.
\end{itemize}
As shown in \cref{tab:target_architecture}, when deriving models for $\mathcal{S}_4$–$\mathcal{S}_6$ from DeiT-Base at pruning rates of 0.60 and 0.40, the \textit{Uniform} strategy consistently achieves the best performance. Considering its superior accuracy and hardware friendliness, we adopt the uniform strategy to determine $e_l$.

{\setlength{\tabcolsep}{4.5pt}
\begin{table}[t]
\caption{Comparison in model accuracy under different features sampling strategies during OFP.}
\label{tab:sampling_strategy}
\begin{center}
\begin{scriptsize}
\renewcommand{\arraystretch}{1.1}
\begin{tabular}{l||cc|cc|cc|cc}
\toprule
\multirow{2}{*}{\textbf{Methods}} &
\multicolumn{2}{c|}{\raisebox{0.5\totalheight}{\textbf{$\mathcal{S}_4$/25}}} & 
\multicolumn{2}{c|}{\raisebox{0.5\totalheight}{\textbf{$\mathcal{S}_5$/25}}} &
\multicolumn{2}{c|}{\raisebox{0.5\totalheight}{\textbf{$\mathcal{S}_6$/25}}} &
\multicolumn{2}{c}{\raisebox{0.5\totalheight}{\textbf{\textit{Avg}}}}\\
\cline{2-9}
& 0.40 & 0.60 & 0.40 & 0.60 & 0.40 & 0.60 & 0.40 & 0.60\\
\midrule
\rowcolor[gray]{0.95}
\textbf{Random} & 94.40 & 89.84 & 97.04 & 92.00 & 96.72 & 92.48 & 96.05 & 91.44\\
\textbf{Max-L2} & 94.16 & 85.36 & 95.68 & 90.48 & 95.92 & 90.24 & 95.25 & 88.69\\
\bottomrule
\end{tabular}
\end{scriptsize}
\end{center}
\end{table}}

\textbf{Sampling Strategy}. During OFP, NuWa requires intermediate features $\mathcal{H}^{\smash{(l)}}\in\mathbb{R}^{\smash{(KN)\times d}}$ from $K$ sampled images to compute the optimal $W_2^{\smash{(l)\prime}}$ according to Eq.~(12). We compare two sampling strategies, i.e., random sampling (Random) and selecting samples with the largest patch-token L2 norms (Max-L2). As shown in \cref{tab:sampling_strategy}, random sampling yields higher model accuracy. 
Moreover, since Random does not involve sorting operations, its $\mathcal{P}_2^{\smash{(2)}}$ in Fig.~9 is smaller than that of Max-L2. Therefore, NuWa adopts random sampling to get calibration features.

\section{Proof}
\label{sec:proof}

In this section, we prove that \cref{eq:prune_mha} and \cref{eq:prune_mlp} are the closed-form solutions to \cref{eq:mha_pruning_problem} and \cref{eq:mlp_pruning_problem}, respectively.

\begin{proof}[\textbf{Proof1: Optimal MHA Pruning}.]
Take $W_Q\in\mathbb{R}^{q\times d}$ and $W_K\in\mathbb{R}^{q\times d}$ ($q<d$) as an example, the optimization objective is to find two matrices $W_Q^\prime\in\mathbb{R}^{q'\times d}$ and $W_K^\prime\in\mathbb{R}^{q'\times d}$ ($q'<q$), such that the Frobenius norm of the difference between $W_{QK}=W_Q^\top W_K\in\mathbb{R}^{d\times d}$ and $W_{QK}^\prime=W_Q^{\prime\top} W_K^\prime\in\mathbb{R}^{d\times d}$ is minimized, i.e.,
\begin{equation}
    \min_{W_Q^\prime,W_K^\prime}\|W_{QK}-W_Q^{\prime\top}W_K^\prime\|_F^2
\end{equation}

We first perform a singular value decomposition (SVD) on $W_{QK}$, obtaining:
\begin{equation}
\begin{aligned}
    & W_{QK} = U_{QK}\Sigma_{QK}V_{QK}^\top,\quad U_{QK},V_{QK}\in\mathbb{R}^{d\times q}\\
    & \Sigma_{QK}=\text{diag}(\sigma_1,\cdots,\sigma_q)\in\mathbb{R}^{q\times q}\\
    & \sigma_1\ge\cdots\ge\sigma_q\ge 0,\quad \text{rank}(W_{QK})\le q
\end{aligned}
\end{equation}

For any $W\in\mathbb{R}^{d\times d}$ with $\text{rank}(W)\le q'$, let $Y=U_{QK}^\top W V_{QK}\in\mathbb{R}^{q\times q}$. Since the Frobenius norm $\|\cdot\|_F$ is invariant under left and right multiplication by orthogonal matrices, we have:
\begin{equation}
\begin{aligned}
    & \|W_{QK}-W\|_F=\|\Sigma_{QK}-Y\|_F\\
    & \text{rank}(Y)=\text{rank}(W)\le q'
\end{aligned}
\end{equation}
Therefore, it suffices to minimize $\|\Sigma_{QK}-Y\|_F^2$ over all matrices $Y$ with $\text{rank}(Y)\le q'$. Expanding the above expression, we obtain:
\begin{equation}
\begin{aligned}
    \|\Sigma_{QK}-Y\|_F^2 & =\|\Sigma_{QK}\|_F^2+\|Y\|_F^2-2\langle \Sigma_{QK},Y\rangle\\
    & = \sum_{i=1}^q\sigma_i^2+\sum_{i}s_i(Y)^2-2\text{tr}(\Sigma_{QK}^\top Y)
\end{aligned}
\end{equation}
where $s_i(Y)$ denotes the $i$-th singular value of $Y$. Since the von Neumann trace inequality satisfies:
\begin{equation}
\text{tr}(\Sigma_{QK}^\top Y)\le \sum_i\sigma_is_i(Y)
\end{equation}
and equality holds when $Y$ and $\Sigma_{QK}$ share the same left and right singular vectors with their singular values aligned in the same order, we have:
\begin{equation}
\begin{aligned}
    \|\Sigma_{QK}-Y\|_F^2 & \ge \sum_{i=1}^q\sigma_i^2+\sum_{i}s_i(Y)^2-2\sum_i\sigma_is_i(Y)\\
    &=\sum_{i=1}^q(\sigma_i-s_i(Y))^2
\end{aligned}
\end{equation}

Since $\text{rank}(Y) \le q'$ implies $s_i(Y) = 0$ for all $i > q'$, it follows that:
\begin{equation}
\begin{aligned}
    \|\Sigma_{QK}-Y\|_F^2 & \ge \sum_{i=1}^{q'}(\sigma_i-s_i(Y))^2+\sum_{i=q'+1}^q\sigma_i^2\\
    & \ge \sum_{i=q'+1}^q\sigma_i^2
\end{aligned}
\end{equation}
The lower bound is achieved when $s_i(Y) = \sigma_i$ for $i \le q'$ and $Y$ shares the same singular vectors as $\Sigma_{QK}$. That is:
\begin{equation}
\begin{aligned}
    & Y^\star = 
    \begin{bmatrix}
    \Sigma_{QK,q'} & 0 \\
    0 & 0
    \end{bmatrix},\quad W^\star = U_{QK}Y^\star V_{QK}^\top\\
    & W_{QK}^{\prime\star}=W^\star=U_{QK}[:,:q']\Sigma_{QK}[:q',:q']V_{QK}[:,:q']^\top
\end{aligned}
\end{equation}
where $\Sigma_{QK,q'}=\text{diag}(\sigma_1,\cdots,\sigma_{q'})$. Therefore, when $W_Q'$ and $W_K'$ are given by:
\begin{equation}
\begin{aligned}
    & W_Q'= (U_{QK}[:,:q']\Sigma_{QK}[:q',:q'])^\top\\
    & W_K' = V_{QK}[:,:q']^\top
\end{aligned}
\end{equation}
the quantity $\|W_{QK} - W_Q^{\prime\top}W_K^\prime\|_F^2$ attains its minimum value 
$\sum_{i = q' + 1}^{q} \sigma_i^2$.

\end{proof}

\begin{proof}[\textbf{Proof2: Optimal MLP Pruning}.]
For given $W_2\in\mathbb{R}^{d\times e}$, $\mathcal{H}\in\mathbb{R}^{(KN)\times e}$, and $\mathcal{I}_r\in\mathbb{R}^{e'}$ ($e'<e$), consider the following problem:
\begin{equation}
\begin{aligned}
    \min_{W_2^{\prime}}\|\mathcal{H}W_2^{\top}-\mathcal{H}[\mathcal{I}_r]W_2^{\prime\top}\|^2_F
\end{aligned}
\end{equation}
which essentially amounts to finding the least-squares solution of $W_2'\in\mathbb{R}^{d\times e'}$.

Write the column vectors of $B=\mathcal{H}W_2^{\top}\in\mathbb{R}^{(KN)\times d}$ as $B = [b_1, \ldots, b_d]$ ($b_i\in\mathbb{R}^{KN}$) and those of $W_2^{\prime\top}\in\mathbb{R}^{e'\times d}$ as $W_2^{\prime\top} = [w_1, \ldots, w_d]$ ($w_i\in\mathbb{R}^{e'}$). Then,
\begin{equation}
\begin{aligned}
    \|\mathcal{H}W_2^{\top}-\mathcal{H}[\mathcal{I}_r]W_2^{\prime\top}\|^2_F= \|\mathcal{H}_rW_2^{\prime\top}-B\|^2_F\\
    = \sum_{i=1}^d\|\mathcal{H}_rw_i-b_i\|^2_2\quad\quad\quad\quad
\end{aligned}
\end{equation}
where $\mathcal{H}_r=\mathcal{H}[\mathcal{I}_r]\in\mathbb{R}^{KN\times e'}$. Hence, the problem decomposes completely column-wise. For any fixed $i$, we aim to minimize $\|\mathcal{H}_rw_i - b_i\|_2^2$. Let $\mathcal{C} = \operatorname{col}(\mathcal{H}_r) \subset \mathbb{R}^{K N}$. According to the orthogonal projection theorem, each $b_i$ admits a unique decomposition:
\begin{equation}
b_i = \underbrace{P_{\mathcal{C}} b_i}_{\in \mathcal{C}} + \underbrace{r_i}_{\perp \mathcal{C}}
\end{equation}
where $P_{\mathcal{C}}$ denotes the orthogonal projection onto $\mathcal{C}$ and $r_i$ is residual vector. For any $w\in\mathbb{R}^{e'}$, we have:
\begin{equation}
\begin{aligned}
    \|\mathcal{H}_rw-b_i\|^2_2&=\|\mathcal{H}_rw_i-P_{\mathcal{C}} b_i+r_i\|_2^2\\
    &=\|\mathcal{H}_rw-P_{\mathcal{C}} b_i\|_2^2+\|r_i\|_2^2\ge\|r_i\|_2^2
\end{aligned}
\end{equation}
This shows that $\min_w \|\mathcal{H}_rw - b_i\|_2^2$ achieves its minimum when $b_i$ is projected onto $\mathcal{C}$, and the image of the optimal solution $w$ must be $\mathcal{H}_rw = P_{\mathcal{C}} b_i$.

By combining the columns together, we obtain the overall optimality condition:
\begin{equation}
\mathcal{H}_rW^\star = P_{\mathcal{C}} B
\end{equation}
Geometrically, this means that each column of $B$ is simultaneously projected onto $\operatorname{col}(\mathcal{H}_r)$. From the necessary and sufficient condition of orthogonal projection, we obtain:
\begin{equation}
\mathcal{H}_r^\top(\mathcal{H}_rW^\star-B)=0
\end{equation}
which is precisely the normal equation for the matrix least-squares problem. It is equivalent to saying that $\mathcal{H}_r W^\star$ is the orthogonal projection of $B$ onto $\operatorname{col}(\mathcal{H}_r)$.

Let $\mathcal{H}_r^{\dagger}$ denote the Moore–Penrose pseudoinverse of $\mathcal{H}_r$. The orthogonal projection operator can then be written as $P_{\mathcal{C}} = \mathcal{H}_r\mathcal{H}_r^{\dagger}$. Hence,
\begin{equation}
\begin{aligned}
    &\mathcal{H}_rW^\star = (\mathcal{H}_r\mathcal{H}_r^{\dagger})B\\
    &\Longrightarrow\quad W^\star=\mathcal{H}_r^{\dagger}B+Z,\quad \mathcal{H}_rZ=0
\end{aligned}
\end{equation}
Here, $Z$ is arbitrary (it does not affect $\mathcal{H}_rW^\star$, and thus yields the same optimality).  
In particular, the minimum-norm optimal solution is:
\begin{equation}
\begin{aligned}
W_2^{\prime\star\top} & =W^\star=\mathcal{H}_r^{\dagger}B=\mathcal{H}_r^\dagger\mathcal{H}W_2^{\top}\\
W_2^{\prime\star} & =W_2\mathcal{H}^\top\mathcal{H}_r^{\dagger\top}=W_2\mathcal{H}^\top((\mathcal{H}_r^\top\mathcal{H}_r)^\dagger \mathcal{H}_r^\top)^\top\\
& = W_2\mathcal{H}^\top\mathcal{H}_r(\mathcal{H}_r^\top\mathcal{H}_r)^\dagger
\end{aligned}
\end{equation}
The minimum objective value is determined by the projection residual:
\begin{equation}
\begin{aligned}
&\min_{W}\|\mathcal{H}_rW-B\|^2_F=\|(I-\mathcal{H}_r\mathcal{H}_r^\top)B\|_F^2\\
&\quad\quad\quad\quad=\sum_{i=1}^d\|(I-\mathcal{H}_r\mathcal{H}_r^\top)b_i\|_2^2
\end{aligned}
\end{equation}
That is, it represents the sum of squared components of $B$ lying in the orthogonal complement of $\operatorname{col}(\mathcal{H}_r)$.
\end{proof}

\section{Implementation Details}
\label{sec:implementation_details}

\begin{table}[t]
    \centering
    \caption{Hyperparameters for the class-specific model derivation process of NuWa. Here, $\alpha\in(0,1)$ denotes the target pruning rate.}
    \begin{tabular}{l||c}
        \toprule
        \textbf{Hyperparameter} & \textbf{Value}\\
        \hline\hline
        \multicolumn{2}{l}{\textit{\textcolor{gray!70}{\textbf{Self-Knowledge Purification}}}}\\
        \rowcolor[gray]{0.95}
        \textbf{Steps} & 10000\\
        \hline
        \textbf{Optimizer} & AdamW~\citep{loshchilov2017decoupled}\\
        \hline
        \rowcolor[gray]{0.95}
        \textbf{Batch Size} & 1\\
        \textbf{Learning Rate (LR)} & LR$_\mathcal{B}$=1e-1, LR$_\mathcal{M}$=1e-3\\
        \hline
        \rowcolor[gray]{0.95}
        \textbf{LR Scheuler} & Constant\\
        \hline
        \textbf{Weight Decay} & 0.05\\
        \hline\hline
        \multicolumn{2}{l}{\textit{\textcolor{gray!70}{\textbf{Optimization-based Fast Pruning}}}}\\
        \multirow{2}{*}{\textbf{Retained Energy Ratio $\rho$}} & \multirow{2}{*}{
              $\begin{array}{c}
              -0.41\alpha^3+0.14\alpha^2 \\
              -0.03\alpha+1.0
              \end{array}$
            }\\
        &\\
        \hline
        \rowcolor[gray]{0.95}
        \textbf{\#Calibration Samples} & 128\\
        \bottomrule
    \end{tabular}
    \label{tab:hyperparam}
\end{table}

\begin{algorithm}[t]
    \caption{Block-Uniform Pruning}
    \label{alg:precise_intermediate_size}
    \begin{algorithmic}[1]
        \renewcommand{\algorithmicrequire}{\textbf{Input:}}
        \renewcommand{\algorithmicensure}{\textbf{Output:}}
        \renewcommand{\algorithmiccomment}[1]{\textcolor{upred}{\hspace{0.0in}\(\triangleright\)~#1}}
        \REQUIRE  neuron counts $\{e_l\}_{l=1}^L$, prune neuron count $e_\text{prune}$
        \vskip 0.05in
        \ENSURE intermediate size list $\{e_l^\prime\}_{l=1}^L$
        \vskip 0.1in
        \STATE $N_\text{total} \gets \sum_{l=1}^Le_l$ \COMMENT{total neurons count}
        \STATE $N_\text{target}\gets\lfloor (N_\text{total} - e_\text{prune}) / L \rfloor$ \COMMENT{target neurons count}
        \STATE $I \gets \mathrm{argsort}(e_l)$ \COMMENT{Sort blocks in ascending order}
        \STATE $e_l^\prime\gets e_l$ for all $l$, $L_\text{count} \leftarrow 0$
        \FOR{each $i$ in $I$}
            \STATE $L_\text{count} \leftarrow L_\text{count} + 1$
            \IF{$e_i \ge N_\text{target}$}
                \STATE $e_i^\prime \leftarrow N_\text{target}$ \COMMENT{\cref{eq:block_uniform_mlp_pruning}}
                \STATE $e_\text{prune} \leftarrow e_\text{prune} - (e_i - e_i^\prime)$
            \ENDIF
            \STATE $N_\text{total} \gets N_\text{total} - e_i$
            \STATE $N_\text{target} \leftarrow \lfloor (N_\text{total} - e_\text{prune}) / (L - L_\text{count}) \rfloor$
        \ENDFOR
        \RETURN $\{e_l^\prime\}_{l=1}^L$
    \end{algorithmic}
\end{algorithm}

We summarize the hyperparameter settings of NuWa in \cref{tab:hyperparam}. In practical implementation, NuWa further optimizes \cref{eq:block_uniform_mlp_pruning} to strictly enforce:
\begin{equation}
\sum_{l=1}^Le_l^\prime=\left(\sum_{l=1}^Le_l\right)-e_\text{prune}
\end{equation}
as described in \cref{alg:precise_intermediate_size}. 

For the Swin Transformer~\cite{liu2021swin}, since the number of input patch tokens varies with image size, the corresponding GFLOPs are not fixed. Therefore, NuWa adopts the number of parameters (\#Param) as the metric to measure the resource constraint and pruning rate. NuWa distributes the total prunable parameter budget proportionally among the MLP modules of each stage in $\mathcal{V}_A$, computes $e_\text{prune}$ for each stage, and determines $e_l'$ for each block following \cref{alg:precise_intermediate_size}.

Moreover, NuWa ensures that the pruned dimensions $q_l'$, $v_l'$, and $e_l'$ are multiples of 8, in order to achieve accelerated inference across a wide range of edge devices.

\section{Baselines}
\label{sec:baseline}

NuWa is compared with seven baselines implemented based on the open-source code from GitHub.
\begin{itemize}[leftmargin=0.2in, topsep=0.01in, itemsep=0.0in]
    \item \textbf{Magnitude Structured Pruning}~\cite{han2015learning}. Given the overall pruning rate $\alpha$, uniform pruning is applied to the $q_l$, $v_l$, and $e_l$ dimensions in each block. Specifically, dimensions with the smallest L2 norms are pruned by removing the corresponding rows or columns in the weight matrices without post-pruning retraining.
    \item \textbf{Wanda-sp}~\cite{sun2023simple}. Wanda-SP is the structured-pruning variant of Wanda. Given the overall pruning rate $\alpha$, it uniformly prunes the $q_l$, $v_l$, and $e_l$ dimensions in each block. Considering the massive activations in Transformer forward propagation, Wanda computes importance scores as the product of activation magnitudes and corresponding weight magnitudes. Weights with the lowest scores are pruned by removing the associated rows or columns in the weight matrices. No retraining is performed after pruning.
    \item \textbf{Numerical Pruning}~\cite{shen2025numerical}. Numerical Pruning estimates the importance of attention heads in MHA and neuron groups in MLP using Newton method. According to the computed importance scores and the overall pruning rate $\alpha$, it adaptively prunes $H_l$ and $e_l$ across blocks. After pruning, a compensation weights is calculated for each pruned weight to restore accuracy, without performing any retraining.
    \item \textbf{X-Pruner}~\cite{yu2023x}. X-Pruner applies learnable masks to $H_l$ and $e_l$ in each block, which are optimized via gradient descent with sparsity-inducing regularization added to the loss function. During training, dimensions with smaller mask values are progressively pruned until convergence. After pruning based on the learned sparse masks, X-Pruner performs retraining to recover accuracy.
    \item\textbf{DC-ViT}~\cite{zhang2024dense}. DC-ViT determines pruning candidates based on each block’s recoverability and the overall pruning rate $\alpha$. It then removes entire MHA modules from the selected blocks and randomly prunes the expansion dimensions $e_l$ of the MLP modules to achieve the target sparsity. After pruning, the model is retrained to restore accuracy.
    \item \textbf{RECAP}~\cite{ilhan2024resource}. RECAP alternates between pruning, fine-tuning, and updating to reduce memory usage while preserving accuracy. It estimates weight importance via a Taylor-based criterion and prunes $H_l$ and $e_l$ across blocks accordingly. Only important weights are updated with a Fisher-based mask, achieving substantial memory savings without full-model retraining.
    \item\textbf{MDP}~\citep{sun2025mdp}.MDP jointly prunes multiple dimensions of ViTs, including the embedding dimension ($d$), the number of attention heads ($H_l$), the query–key and value–output dimensions ($q_l,v_l$), and the MLP intermediate dimension ($e_l$). It formulates pruning as a mixed-integer nonlinear program (MINLP) problem under latency budgets, solved with Hessian-based importance scores and a precomputed latency lookup table (LUT). After pruning, the model is retrained to restore accuracy.
\end{itemize}

\begin{figure}[t]
   \centering
   \includegraphics[width=1.0\linewidth]{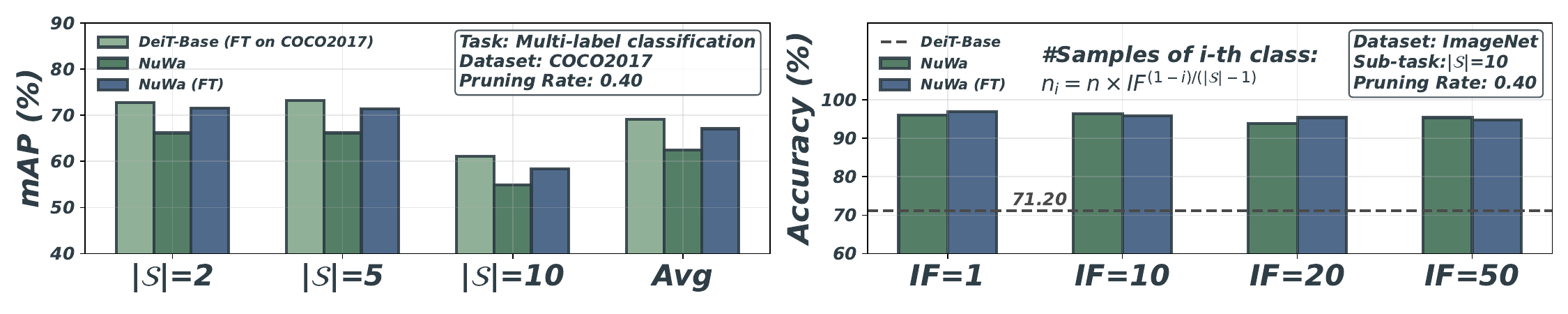}
   \caption{Performance of NuWa under class co-occurrence and long-tailed settings.}
   \label{fig:complex_settings}
\end{figure}

\section{Performance under Complex Settings}
To account for the complexity of real-world deployment scenarios, we further evaluate NuWa under two challenging settings: class co-occurrence and long-tailed distributions.

For the class co-occurrence setting, where multiple categories may appear within a single image, we sample multi-label classification sub-tasks of varying scales from COCO2017 and derive edge ViTs from a DeiT-Base model fine-tuned on COCO2017. As shown in the left panel of \cref{fig:complex_settings}, at a pruning rate of 0.40, the models derived by NuWa achieve an average precision of 97.13\% relative to the base ViT. This result demonstrates the strong generality of NuWa in complex multi-label scenarios.

For the long-tailed setting, we construct long-tailed variants of each ImageNet sub-task dataset using the following transformation:
\begin{equation}
n_i=n\times\text{IF}^{(1-i)/(|\mathcal{S}|-1)}
\end{equation}
where $n$ is the per-class count in the balanced set, $n_i$ is the count for the $i$-th class after imbalance, $|\mathcal{S}|$ is the number of classes in the sub-task, and IF is the imbalance factor (IF=1 means balanced). As shown in the right panel of \cref{fig:complex_settings}, for sub-tasks containing 10 classes, the models derived by NuWa maintain stable accuracy across different imbalance levels at a pruning rate of 0.40, demonstrating its robustness under long-tailed data distributions.

\section{Pseudocode}

\begin{algorithm}[H]
    \caption{NuWa – Class-Specific Model Derivation}
    \label{alg:nuwa}
    \begin{algorithmic}[1]
        \renewcommand{\algorithmicrequire}{\textbf{Input:}}
        \renewcommand{\algorithmicensure}{\textbf{Output:}}
        \renewcommand{\algorithmiccomment}[1]{\textcolor{upred}{\hspace{0.0in}\(\triangleright\)~#1}}
        \REQUIRE  pretrained all-class base ViT $\mathcal{V}_B$, sub-task $\mathcal{S}$, class-specific data $\mathcal{D_S}$, overall pruning rate $\alpha$
        \vskip 0.05in
        \ENSURE lightweight class-specific edge ViT $\mathcal{V}_E$
        \vskip 0.1in
        \STATE \textbf{\# Self-Knowledge Purification (SKP)}
        \STATE $\mathcal{M}=\{M^{(l)}\}_{l=1}^L\gets\{\mathbf{1}\in \mathbb{R}^{e_l}\}_{l=1}^L$
        \STATE $\mathcal{B}=\{\beta^{(l)}\}_{l=1}^L\gets\{5.0\}_{l=1}^L$
        \STATE Freeze parameters of $\mathcal{V}_B$
        \STATE Embed $\mathcal{M}$ and $\mathcal{B}$ into the MLP modules of $\mathcal{V}_B$
        \STATE Train $\mathcal{M}$ and $\mathcal{B}$ on $\mathcal{D_S}$ under supervision of $\mathcal{L}_T$
        \hspace{1.5em}\COMMENT{$\mathcal{M}$ and $\mathcal{B}$ are involved in the forward propagation according to \cref{eq:mask_binarization} and \cref{eq:mlp_binarized}}
        \STATE Prune $\mathcal{V}_B$ according to $\mathcal{M}$ and $\mathcal{B}$ to obtain the anchor model $\mathcal{V}_A$
        \STATE \textbf{\# Optimization-based Fast Pruning}
        \STATE Compute $\rho$ based on $\alpha$ \COMMENT{equation in \cref{tab:hyperparam}}
        \STATE Compute $\{q_l'\}_{l=1}^L$ and $\{v_l'\}_{l=1}^L$ based on $\rho$ \COMMENT{\cref{eq:retained_energy_rate}}
        \STATE $\mathcal{V}_A\gets$ Prune MHA modules of $\mathcal{V}_A$ using SVD based on $\{q_l'\}_{l=1}^L$ and $\{v_l'\}_{l=1}^L$ \COMMENT{\cref{eq:prune_mha}}
        \STATE $e_{\text{prune}} = \left((1-\alpha)\mathcal{F}(\mathcal{V}_B) - \mathcal{F}(\mathcal{V}_A)\right) / 2Nd$ \COMMENT{$\mathcal{F}$ denotes the function that computes GFLOPs or \#Params}
        \STATE Compute $\{e_l'\}_{l=1}^L$ based on $e_{\text{prune}}$ \COMMENT{\cref{eq:block_uniform_mlp_pruning} and \cref{alg:precise_intermediate_size}}
        \STATE $\{a^{(l)}\}_{l=1}^L$,$\{\mathcal{H}^{(l)}\}_{l=1}^L\gets$ Perform one epoch of forward propagation of $\mathcal{V}_A$ on $\mathcal{D_S}$
        \STATE $\mathcal{V}_A\gets$ Prune MLP modules of $\mathcal{V}_A$ based on $\{e_l'\}_{l=1}^L$, $\{a^{(l)}\}_{l=1}^L$, and $\{H^{(l)}\}_{l=1}^L$
    \end{algorithmic}
\end{algorithm}

\section{Limitations and Future Works}
A common limitation of NuWa, as well as other pruning methods that do not rely on post-pruning retraining, is that excessive pruning inevitably leads to significant accuracy degradation. As shown in \cref{fig:training_free_comparison}, pruning-only methods, i.e., Random Pruning, Magnitude Pruning, and Wanda-sp suffer sharp accuracy drops when the pruning rate exceeds 20\%. Numerical Pruning, which introduces compensation matrices to recover accuracy, maintains stable performance until around 50\% pruning. In contrast, NuWa effectively removes class-detrimental weights through SKP, enabling the derived models to outperform the base ViT on target classes even at a pruning rate of 60\%. However, when the pruning rate exceeds 70\%, noticeable accuracy degradation still occurs. Moreover, as the sub-task size increases (e.g., $|\mathcal{S}|>50$), the performance of derived models also declines.

To address the performance drop under extremely high pruning rates or large-scale sub-tasks, we envision two promising directions. First, lightweight retraining can be introduced to recover accuracy, as NuWa-derived models retain more class-specific knowledge and therefore require less retraining overhead. Second, an offline-trained hypernetwork could be developed to predict weight updates, enabling the online derivation process to efficiently map task specifications, such as model architecture, sub-task, and pruning rate, to corresponding pruned weights.
